\documentclass[pdflatex,sn-mathphys-num]{sn-jnl}
\usepackage{graphicx} 
\usepackage{multirow}%
\usepackage{amsmath,amsfonts, amssymb}
\usepackage{amsthm}%
\usepackage{mathrsfs}%
\usepackage[title]{appendix}%
\usepackage{xcolor}%
\usepackage{textcomp}%
\usepackage{manyfoot}%
\usepackage{booktabs}
\usepackage{algorithm}
\usepackage{algpseudocode}
\usepackage{makecell}

\usepackage{listings}%
\usepackage{amsthm}

\usepackage{tikz}
\usetikzlibrary{positioning}

\tikzset{
  info/.style={
    rectangle,
    draw=none,
    font=\tiny,
    align=center
  }
}

\newcommand{\clusterinfo}[4]{%
  \node[info, right=2pt of #1] 
  {$q(#1)=#4$\\$LB=#2$\\$UB=#3$};
}

\begin{document}

\title{FOSC-X: An Extended Framework for Optimal Local Cuts and Non-Horizontal Cluster Selection from Clustering Hierarchies}

\author*[1]{\fnm{Connor} \sur{Simpson}}\email{simpson@imada.sdu.dk}

\author[1]{\fnm{Ricardo J. G. B.} \sur{Campello}}\email{campello@imada.sdu.dk}

\affil[1]{\orgdiv{Department of Mathematics and Computer Science}, \orgname{University of Southern Denmark}, \orgaddress{\city{Odense}, \country{Denmark}}}

\abstract{Hierarchical clustering represents data as a cluster tree capturing nested structure at multiple resolutions. Extracting a flat clustering solution from a hierarchy is a common task in practical cluster analysis and can be formulated as an optimisation problem. Existing approaches focus on finding a single optimal solution. We introduce FOSC-X, a framework for extracting the top-$M$ globally optimal flat clusterings from local, non-horizontal cuts of a hierarchical cluster tree, while optionally enforcing constraints on the number of clusters. This enables automatic identification of multiple high-quality alternative clusterings that capture different aspects of the hierarchical structure.
Without constraints, the top-$M$ problem can be solved in polynomial time using dynamic programming, exploiting the property that locally optimal partial candidates within subtrees can be combined to form globally optimal solutions while automatically determining the number of clusters. However, this can lead to solutions with numbers of clusters that are ultimately undesirable --- e.g., too large to be meaningful or practically analysed within a particular application domain. Imposing cluster-count constraints breaks the optimality property underlying the unconstrained dynamic programming approach, since locally optimal partial candidates may no longer combine into feasible globally optimal solutions. 
FOSC-X addresses this challenge through a dynamic programming strategy that maintains compact sets of feasible candidates using lower and upper feasibility bounds while pruning infeasible or dominated combinations. The resulting method guarantees optimal rankings of the top-$M$ solutions with linear-time complexity in the number of cluster nodes and dataset size, both with and without cluster-count constraints. Experiments show that FOSC-X efficiently reveals alternative clustering structures overlooked by single-solution extraction methods.}


\maketitle

\section{Introduction} \label{sec:intro}

Clustering is a fundamental task in unsupervised learning, aiming to group observations whose members are somehow more similar or related to each other than to other observations within the data. Unlike supervised learning problems, clustering typically lacks ground-truth labels, and the notion of a ``correct'' clustering is subjective. This notion is often ambiguous even under objective definitions of the clustering problem. For instance, multiple valid clusterings of the same dataset may coexist depending on the granularity level (resolution) or the intended downstream application \cite{JAIN2010651}. As a result, cluster analysis is inherently exploratory and selecting an appropriate clustering solution often requires considering multiple plausible candidates \cite{VonLuxburg2012,HENNIG201553}.

Hierarchical clustering provides a natural multi-resolution framework in which nested groupings of observations are represented as a cluster tree rather than a single flat clustering \cite{Jain1988,everitt2001cluster}. A common class of methods are agglomerative clustering algorithms, such as Single-Linkage, Complete-Linkage, Average-Linkage, and Ward-Linkage, which begin with each observation as its own cluster and iteratively merge the most similar clusters according to a linkage criterion \cite{LeonardKaufman1990}. Hierarchical structures can also arise from density-based approaches; for example, HDBSCAN* constructs a hierarchy by determining how clusters form and split across varying density levels following the statistical notion of density-contour trees \cite{campello2013density, campello2015hierarchical}. This multi-scale representation is widely used in exploratory data analysis, as it enables analysts to examine both coarse global structure and fine-grained subdivisions within the data \cite{EZUGWU2022104743, Ran_Xi_Lu_Wang_Lu_2022}. Such hierarchical organisation is particularly valuable in domains where observations exhibit structured similarity relationships, including biological taxonomy and phylogenetic analysis \cite{Segal2002,Eisen98}, as well as gene expression and cancer subtype analysis where biologically meaningful subgroups may exist at multiple levels of granularity \cite{Han_Zhang_Gao_Bu_2024}.

In practice, hierarchical clustering is often used as an intermediate representation from which a single flat clustering is extracted for interpretation or downstream tasks \cite{Nielsen2016}. Such a clustering is obtained by selecting a set of disjoint clusters from the hierarchy, potentially leaving some observations unassigned as noise, as in density-based approaches such as HDBSCAN*. Traditional approaches rely on heuristic rules, such as cutting the tree at a fixed height, applying stopping criteria based on linkage statistics, or selecting a partition using internal validity indices \cite{Milligan1985a,Hennig_Meila_Murtagh_Rocci_2015}. These methods correspond to making a single global decision on the tree corresponding to a horizontal cut across a single level of the tree. They therefore cannot easily adapt to heterogeneous cluster scales, varying densities, or complex hierarchical structure \cite{campello2013fosc}.

Optimisation-based approaches instead formulate cluster extraction as the problem of directly optimising an objective function --- as a quantitative measure of clustering quality --- over the hierarchy. A prominent example is the Framework for Optimal Selection of Clusters (FOSC) \cite{campello2013fosc}, most notably used in the HDBSCAN* algorithm, which selects the set of disjoint clusters that maximises a given clustering quality measure without restricting solutions to a single horizontal cut. FOSC performs this optimisation efficiently using a bottom-up dynamic programming strategy in which locally optimal subtree solutions combine to form a globally optimal clustering. A limitation of FOSC is that it returns only a single globally optimal clustering, providing limited flexibility for exploring alternative high-quality solutions. 
Also, it offers no mechanism to control clustering granularity in the resulting single solution.

Other existing extraction approaches also remain limited in one or more of the following important aspects \cite{marti2017cut, Ge03042026, boucherie2025adaptivecutrevealsmultiscale}: (i) like FOSC, they do not allow retrieval of a user-defined number ($M > 1$) of top-$M$ solutions that are globally optimal and directly comparable according to the quality measure adopted for optimisation, regardless of the number of clusters they contain; (ii) they may not provide the user with full control over the level of flexibility with which the number of clusters is handled, namely, the choice between setting a fixed count, $k$, or instead constraining the number of clusters within an interval, [$k_{\min},k_{\max}$], or alternatively allowing for the number of clusters to be automatically determined in a completely unconstrained fashion; (iii) they may offer little to no flexibility in terms of clustering quality measures to be optimised; and/or (iv) they may deploy heuristic approximations with no optimality guarantees.

To address these limitations, we introduce an extended framework for cluster extraction from hierarchies, termed \textit{FOSC-X}\footnote{A Python library for the FOSC-X framework is available at \url{https://github.com/Campello-Lab/FOSC-X}.}. FOSC-X generalises the original FOSC formulation by enabling the enumeration of multiple high-quality clusterings while supporting flexible constraints on the number of extracted clusters. Rather than requiring the number of clusters to be fixed in advance, FOSC-X allows users to specify lower and upper limits on the desired clustering granularity. The framework computes the top-$M$ highest-scoring flat clusterings under a specified objective function, optionally restricted to solutions whose number of clusters lies within the user-defined range.

While unconstrained top-$M$ extraction follows naturally from the additive structure of the original FOSC strategy, introducing explicit cluster-count constraints fundamentally changes the optimisation problem. In the constrained setting, locally optimal subtree solutions may no longer combine into globally feasible solutions, breaking the optimal substructure property relied upon by FOSC's dynamic programming approach. FOSC-X addresses this challenge through a feasibility-aware dynamic programming strategy with dominance pruning, enabling the exact recovery of the constrained top-$M$ solutions in linear time with respect to the size of the cluster tree.

The main contributions of this work are as follows:

\begin{itemize}
\item We extend the FOSC framework from single-solution extraction to the exact enumeration of the top-$M$ highest-scoring flat clusterings contained within a hierarchical cluster tree.

\item We introduce a new, constrained top-$M$ extraction formulation supporting (optional) explicit lower and upper limits on the number of clusters, $k_{\min}$ and $k_{\max}$. This formulation seamlessly include the case of a fixed number of clusters, $k$, as a particular case (by setting $k_{\min} = k_{\max} = k$). 

\item We develop a feasibility-aware dynamic programming algorithm with dominance pruning that exactly recovers the constrained top-$M$ solutions while scaling linearly with the size of the hierarchy.

\item Through experimental evaluation, we demonstrate that FOSC-X uncovers meaningful alternative cluster structures that are often overlooked by single-solution extraction methods.
\end{itemize}

By extending the cluster extraction problem from single-solution optimisation to constrained multi-solution enumeration, FOSC-X provides a principled framework for exploratory analysis of alternative flat clusterings while preserving the globally optimal solution identified by the original FOSC formulation in the unconstrained case. The framework is agnostic to the underlying hierarchical representation and can be applied to binary or non-binary trees arising e.g. from distance-based or density-based hierarchical clustering methods. This includes condensed hierarchies containing noise observations, such as simplified dendrograms or density-based cluster trees.

The remainder of the paper is organised as follows. Section~\ref{sec:Background} reviews existing optimisation-based cluster extraction methods, as well as quality measures used within the original FOSC framework. Section~\ref{sec:motivation} presents motivating examples illustrating the limitations of existing single-solution extraction approaches and the importance of flexible cluster-count constraints. Section~\ref{sec:FOSC-X} introduces the proposed FOSC-X framework, including the dynamic programming formulation and feasibility-aware pruning strategy. Section~\ref{sec:noise} discusses the treatment of noise and the effects of tree condensation on the FOSC-X framework. Section~\ref{subsec:complexity} presents the computational complexity analysis of FOSC-X. Section~\ref{sec:eval} evaluates the proposed method on a collection of benchmark datasets. Finally, Section~\ref{sec:Conclusion} concludes the paper.

\section{Related Work} \label{sec:Background}

\subsection{Optimisation-based cluster extraction}

Methods have been previously proposed to enable multi-level extraction from hierarchical clusterings, addressing the limitations of traditional single horizontal cut strategies. Rather than enforcing a single global resolution across the cluster tree, these approaches allow clusters to be selected from different depths, typically through the optimisation of cluster quality criteria or other adaptive selection procedures.

A prominent example is the Framework for Optimal Selection of Clusters (FOSC) \cite{campello2013fosc}, which formulates flat clustering extraction as the problem of locally selecting disjoint subsets of candidate clusters from the hierarchy that together maximise an objective function measuring global clustering quality.
FOSC performs cluster extraction using a bottom-up dynamic programming procedure over the hierarchical cluster tree. Beginning at the leaves and traversing upward, the algorithm computes an optimal partial solution for each subtree independently. At each node, two alternatives are compared: selecting the cluster represented by the current node as a single cluster in the final clustering, or instead selecting the combination of optimal solutions obtained from its child subtrees. Conceptually, by recursively choosing the better of these alternatives, FOSC constructs a globally optimal flat clustering from locally optimal subtree solutions in a single traversal of the tree with linear time complexity.
Optimality and efficiency can be simultaneously achieved by assuming that the objective function satisfies two key properties: \emph{locality}, meaning the quality of individual clusters can be evaluated independently, and \emph{additivity}, meaning the quality of a candidate flat clustering solution is the sum of the quality scores of its clusters. Under these assumptions, FOSC returns a single globally optimal solution in which the number of clusters is automatically determined.

Alternative approaches impose different optimisation objectives, strategies, or constraints. DynaCut \cite{marti2017cut, 10.1016/j.ipl.2011.10.006}, for example, computes an optimal flat clustering (also under implicit locality and additivity assumptions) subject to a fixed number of clusters, $k$, by maintaining optimal partial solutions for different cluster counts within each subtree of a binary cluster tree. For any specified value of $k$, the method returns only a single optimal partition. 
The authors report a computational complexity of $O(nk \log n)$, where $n$ denotes the dataset size, although in practice the need to maintain solutions across multiple cluster counts can lead to a substantially higher computational cost, as it will be explored in Section~\ref{sec:empiricalComp}.

Another approach to flat clustering extraction from hierarchies is presented in \cite{Ge03042026}, which adapts the CART tree-pruning framework \cite{Breiman_Friedman_Olshen_Stone_2017} by replacing class-label impurity measures with an unsupervised monotone loss such as within-cluster dispersion. The resulting objective function is effectively a weighted combination of the dispersion loss and a regularisation (penalty) term on the cluster count. Starting from the full hierarchy, the method recursively trims off branches to generate a nested sequence of candidate partitions with progressively fewer clusters, corresponding to increasing values of penalty weights, where the leaf nodes of each resulting pruned tree define the associated disjoint, flat clustering. This formulation imposes stricter requirements on the objective function than dynamic-programming approaches such as FOSC, limiting the range of compatible quality measures. Unlike DynaCut, it does not guarantee that a solution with an arbitrary cluster count $k$ will be produced, since the resulting collection of nested partitions may contain none with $k$ clusters. For each number of clusters effectively represented in the produced nested partitions, there will be no more than a single partition with that particular granularity, except for unlikely cases of numerical ties. Most importantly, the multiple partitions produced, each of which has a unique cluster count, are not comparable as each effectively corresponds to the unique optimal solution for a different optimisation criterion (different dispersion versus granularity regularisation trade-offs). This means that they cannot be ranked, and selecting the best one(s) among them would require a separate clustering quality criterion that is fairly comparable across different numbers of clusters (just like in the classic scenario of global horizontal cuts). However, selection according to such a separate criterion would only be optimal among the relatively small collection of partitions produced by this extraction procedure and subject to that criterion's assessment, \emph{not} over the entire space of candidate solutions from the hierarchy.

Another recent approach is the adaptive cut framework \cite{boucherie2025adaptivecutrevealsmultiscale}, which formulates cluster selection as an optimisation problem over multi-level cuts solved approximately using Markov chain Monte Carlo and simulated annealing. While more powerful than fixed-height cutting heuristics, the approach remains heuristic in essence and approximate in practice, offering no global optimality guarantees.

\subsection{Quality measures} \label{sec:backquality}

In flat clustering, different cluster validity indices capture different aspects of cluster structure, and their suitability may depend on the characteristics of the data and the intended purpose of the analysis \cite{Vendramin2010,SimpsonCampelloStojanovski2026,Simpson2026}. Similarly, optimisation-based cluster extraction frameworks, such as FOSC and FOSC-X, may benefit from different objective functions depending on the application. Several quality measures satisfying the FOSC requirements of additivity and locality have been proposed in the literature.
The most common is \emph{cluster stability} \cite{campello2013fosc}, which evaluates how persistently a cluster exists across levels of the hierarchy. In density-based settings, this corresponds to the \emph{Excess of Mass (EOM)} criterion \cite{campello2013density, campello2015hierarchical}, used in HDBSCAN*. Other measures instead operate on the data representation, such as graph-based objectives including \emph{Modularity $Q$}  \cite{10.1007/978-981-13-6661-1_20} or Partition-Free Cluster Evaluation (PFCE) \cite{thompson2025pfce}. The choice of quality measure is application dependent and influences the structure of the extracted clustering.

In addition to these unsupervised clustering quality measures, several semi-supervised measures have been developed. The original paper where FOSC was introduced \cite{campello2013fosc} supplied a semi-supervised alternative based on must-link and cannot-link constraints. Subsequently, in \cite{Gertrudes_Zimek_Sander_Campello_2019}, label-based semi-supervised measures were also introduced based on $B^3$  precision and recall. 

FOSC-X remains agnostic to the choice of quality measure provided the corresponding objective function satisfies the same locality and additivity assumptions required by the original FOSC, making it compatible with the existing measures developed for cluster extraction using that framework.

\section{Motivating Example} \label{sec:motivation}

The following examples 
illustrate two key challenges in cluster extraction from hierarchies that motivate the \mbox{FOSC-X} framework. First, multiple high-quality flat clusterings may coexist within the same hierarchy, with near-optimal candidates possibly providing valid alternative interpretations of the data that may also be useful and potentially even more meaningful to the user as compared to the top-1 globally optimal solution according to the quality measure adopted for extraction. Second, there are scenarios where unconstrained optimisation objective functions may tend to favour either coarse-grained or highly fragmented flat clustering solutions, motivating the incorporation of explicit cluster-count constraints.

\begin{figure}[!t]
	\centering
	\includegraphics[width=0.8\textwidth, clip]{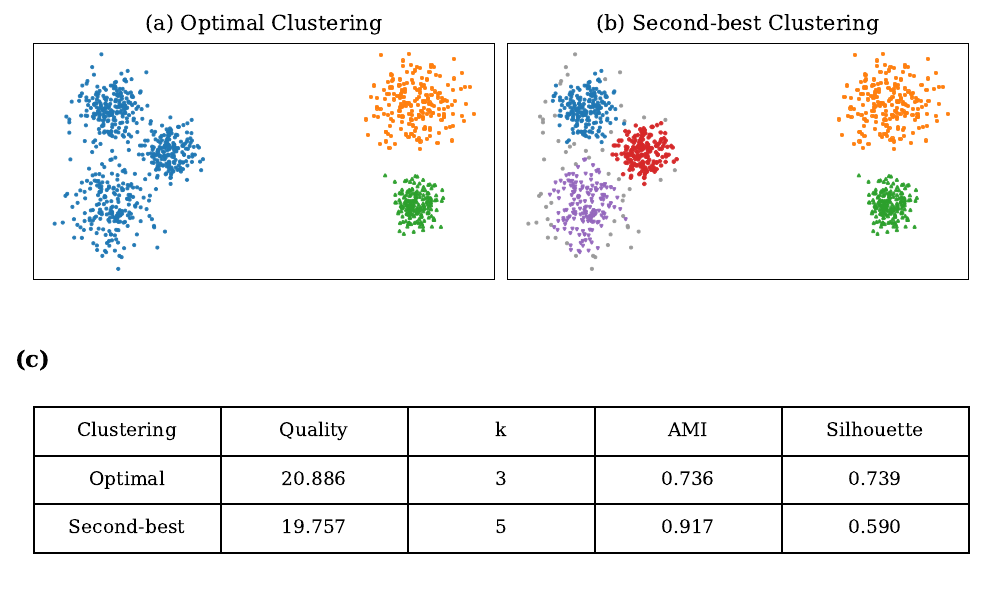} 
	\caption{Five synthetically generated ground-truth clusters and the top-2 partitions extracted from an HDBSCAN* hierarchy. (a) The optimal partition extracted using FOSC (or unconstrained \mbox{FOSC-X}). (b) The second-best partition identified by unconstrained FOSC-X. (c) Comparison of clustering quality measures for both solutions; in the second row, ``Quality'' stands for Excess of Mass (EOM), which is used by FOSC/FOSC-X in this example, and it is also the default measure in HDBSCAN*.}
	\label{fig:Toy1}
\end{figure}

Figure~\ref{fig:Toy1} illustrates a limitation of restricting extraction to a single optimal solution using a toy dataset obtained from \cite{ClusteringDatasets}. Although the globally optimal FOSC clustering achieves the highest objective function value and internal validity (Silhouette \cite{Rousseeuw1987}), a near-optimal alternative more closely reflects the underlying generating process, recovering all five ground-truth clusters and achieving substantially higher external agreement (Adjusted Mutual Information --- AMI \cite{JMLR:v11:vinh10a}). Both solutions offer alternative, valid ``views'' of the data from different scales, yet only the first is detected by the original FOSC. This illustrates that the objective-optimal clustering is not always the only (or necessarily the most) semantically meaningful flat interpretation of a hierarchy, motivating the exploration of multiple high-quality solutions.

\begin{figure}[!t]
	\centering
	\includegraphics[width=\textwidth, clip]{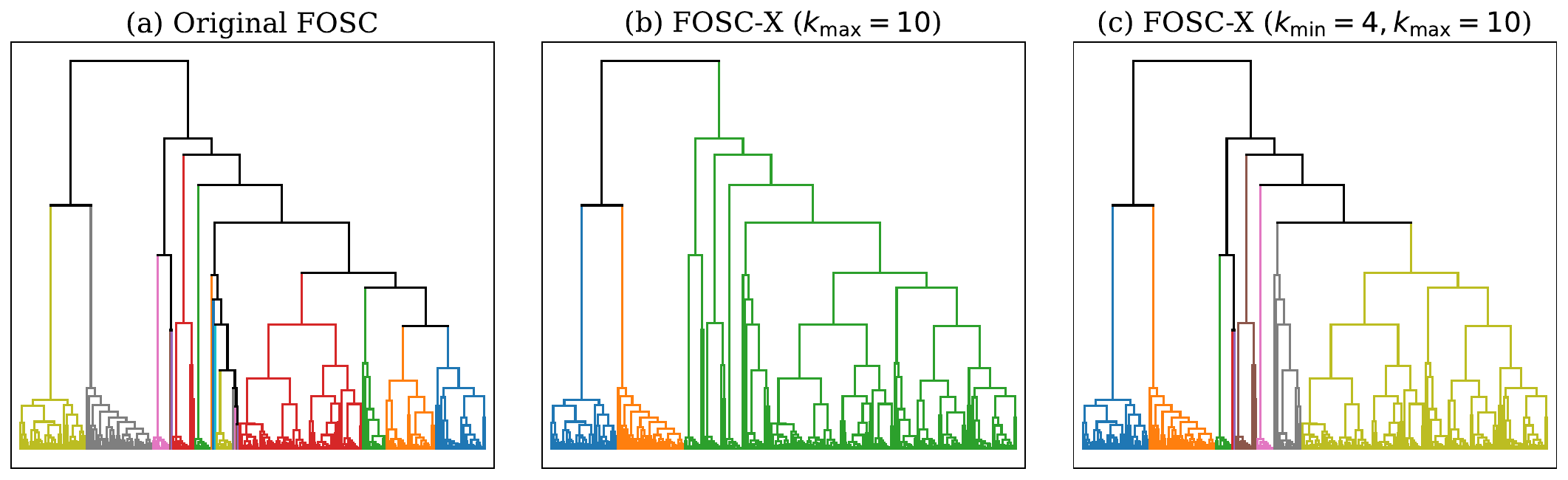}  
	\caption{Flat clustering extraction results on a dendrogram produced from a 2D synthetic dataset containing 280 observations. The dendrogram was generated using Average-Linkage with Euclidean distance. (a) Original FOSC (equivalently, unconstrained \mbox{FOSC-X}) exhibiting a highly fragmented, 17-cluster selection. (b) Constrained \mbox{FOSC-X} with $k_{\max}=10$, producing a three-cluster solution selected via non-horizontal cuts. (c) Constrained \mbox{FOSC-X} with $k_{\min}=4$ and $k_{\max}=10$, producing a nine-cluster solution. All methods use Stability as the objective function for extraction.} 
	\label{fig:Toy2}
\end{figure}

Figure~\ref{fig:Toy2} illustrates a case where unconstrained extraction by the original FOSC (panel (a)) produces a highly fragmented flat clustering with 17 clusters, including some outlying singletons. While such a result may represent a valid interpretation of the data as the corresponding coloured dendrogram visually suggests, 
such an interpretation may not be suitable in applications where the user or domain expert favours broader groupings over highly fine-grained structure consisting of many smaller sub-clusters.
By introducing an upper limit on the number of clusters ($k_{\max} = 10$ in Figure~\ref{fig:Toy2}, panel (b)), \mbox{FOSC-X} instead extracts an alternative three-cluster solution that preserves a coarser and potentially more interpretable structure. Notably, this solution cannot be recovered through a global horizontal cut of the hierarchy, instead requiring local cuts at different levels of the tree. Furthermore, this solution is unlikely to be identified through inspection of top-ranked alternatives alone, as it appears only as the 19th-best unconstrained solution and would require examining a large number of candidates.
Additionally introducing a lower limit on the number of clusters ($k_{\min} = 4$  in Figure~\ref{fig:Toy2}, panel (c)), \mbox{FOSC-X} extracts an alternative solution in which the stable cluster structure contained within the right-hand branch of the dendrogram is retained, while separating the more weakly supported, 
smaller-cluster branches. 
This illustrates how cluster-count constraints can guide extraction within an acceptable range of granularity levels without requiring the user to specify an exact number $k$ of clusters (which is still possible by setting $k_{\min} = k_{\max} = k$). 

Together, these examples motivate the FOSC-X framework developed in the following section, which jointly supports multi-solution (top-$M$) extraction and flexible cluster-count constraints.

\section{FOSC-X} \label{sec:FOSC-X}

We now introduce the FOSC-X framework and derive the corresponding dynamic programming formulation for top-$M$ flat cluster extraction from hierarchies under optional cluster-count constraints. 
We first introduce the notation and derive the recursive definition underpinning unconstrained top-$M$ extraction. We then extend the formulation to support minimum and maximum cluster-count constraints before presenting the associated pruning strategy and computational complexity analysis.

\subsection{Efficient unconstrained selection of top-$M$ solutions}  \label{sec:topm}

The first objective of FOSC-X is to extend the original FOSC framework 
so it can compute the top-$M$ globally optimal flat clusterings within a hierarchy for an arbitrary number $M \geq 1$ chosen by the user. Although the number of possible clusterings grows exponentially with the size of the tree, this can still be achieved efficiently using a modification of the original bottom-up dynamic program. In particular, we show within this section that retaining the top-$M$ partial optimal solutions locally at each node suffices to exactly recover the top-$M$ globally optimal solutions at the root.

Let $\mathcal{T} = \{C_0, C_1, \dots, C_N\}$ be a rooted tree with $N$ nodes, where $C_0$ is the root. Each node $C_i \in \mathcal{T}$ represents a candidate cluster with an associated quality score $q(C_i) \in \mathbb{R}$. For each node $C_i$, we denote its set of children by $\mathrm{ch}(C_i)$.

A \emph{valid clustering} of the subtree rooted at a node $C_i$ (included) is defined as a set of nodes such that exactly one node is selected along every root-to-leaf path in that subtree. We denote such a clustering 
by $z_{C_i} \subseteq \mathcal{T}$. Its total quality is given by the aggregated sum of the individual quality scores of the nodes it contains:
\begin{equation}
s_{C_i} := q(z_{C_i}) = \sum_{C' \in z_{C_i}} q(C').
\end{equation}

For each node $C_i$, let $\mathcal{Z}(C_i)$ denote the set of all possible valid clusterings of the subtree rooted at $C_i$, i.e., the space of candidate partial solutions that  are locally considered by the dynamic program at $C_i$'s subtree. The corresponding set of achievable scores is
\begin{equation}
S(C_i) = \{ s_{C_i} \mid z_{C_i} \in \mathcal{Z}(C_i) \},
\end{equation}
where each clustering $z_{C_i} \in \mathcal{Z}(C_i)$ has an associated objective value $s_{C_i} \in S(C_i)$. 

We now describe a bottom-up dynamic programming approach to compute the optimal clusterings. Although clusterings are the underlying objects, the formulation can be expressed in terms of their scores. In particular, we operate on the sets of achievable scores $S(C_i)$, while maintaining implicit associations to the corresponding clusterings $z_{C_i} \in \mathcal{Z}(C_i)$.\footnote{This means that each element in a set of scores is implicitly a pair of a score and its associated unique clustering, which means there may be more than one element sharing the same score in the set.}
Formally, we define
\begin{equation}
S^{topM}(C_i) = \mathrm{TopM}(S(C_i)),
\end{equation}
where $\mathrm{TopM}(A)$ returns the $M$ largest elements of a finite set $A$ (or all elements if $|A| < M$).
Then, beginning at the leaves of the tree, at each leaf node the only valid clustering at that node's subtree consists of the corresponding leaf cluster itself:
\begin{equation}
S^{topM}(C_i) = \{ q(C_i) \}.
\end{equation}

If $C_i$ is an internal node, there are two alternatives. First, $C_i$ may be treated as a single cluster, yielding score $q(C_i)$. Second, we may expand the node and combine partial clusterings from its children. In this case, a combined clustering is formed by selecting one clustering $z_{C'} \in \mathcal{Z}(C')$ for each child $C' \in \mathrm{ch}(C_i)$ and taking their union,
\begin{equation}
z_{C_i} = \bigcup_{C' \in \mathrm{ch}(C_i)} z_{C'}.
\end{equation}
When considering the set of all $M^{|\mathrm{ch}(C_i)|}$ possible such combinations of the top-$M$ partial clusterings from each child of $C_i$, their scores are obtained by adding the corresponding score from each child, i.e., the collection of the resulting combined scores is the sumset (a.k.a. Minkowski sum) of the top-$M$ scores from each child of $C_i$:
\begin{equation}
\bigoplus_{C' \in \mathrm{ch}(C_i)} S^{topM}(C')
=
\left\{
\sum_{C' \in \mathrm{ch}(C_i)} s_{C'}
\;\middle|\;
s_{C'} \in S^{topM}(C')
\right\}.
\label{eq:sum}
\end{equation}

We then take the union of the two alternative options mentioned above (i.e., expanding $C_i$ or not) and retain only the top-$M$ values:
\begin{equation}
S^{topM}(C_i)
=
\mathrm{TopM}\!\left(
\{ q(C_i) \}
\;\cup\;
\bigoplus_{C' \in \mathrm{ch}(C_i)} S^{topM}(C')
\right).
\label{eq:recursion}
\end{equation}

The final output at the root, $S^{topM}(C_0)$, contains the top-$M$ achievable clustering scores for the full tree.

\subsection{Correctness}

The recursive definition can be materialised efficiently in an iterative bottom-up fashion because each set \(S^{\mathrm{top}M}(C_i)\) stores at most \(M\) values. This truncation is valid since the objective is additive: for a node \(C_i\) with children \(C_a\) and \(C_b\), the Minkowski sum
\begin{equation}
    S^{\mathrm{top}M}(C_a)\oplus S^{\mathrm{top}M}(C_b)
\end{equation}
contains all pairwise sums \(s_a+s_b\) where $s_a \in S^{\mathrm{top}M}(C_a)$ and $s_b \in S^{\mathrm{top}M}(C_b)$. If \(s_a' \le s_a\), then for any fixed \(s_b\),
\begin{equation}
s_a' + s_b \le s_a + s_b,
\end{equation}
so replacing a local candidate by another with a smaller local score cannot increase the total score. If \(s_a' \notin S^{\mathrm{top}M}(C_a)\), then by definition there exist at least \(M\) scores in \(S^{\mathrm{top}M}(C_a)\) that are greater than or equal to \(s_a'\). Every solution involving \(s_a'\) is therefore dominated by at least \(M\) solutions formed using scores from \(S^{\mathrm{top}M}(C_a)\), implying that no such solution can appear among the top-\(M\) combinations at \(C_i\). The same argument applies to all children in case of non-binary trees.

Correctness follows by induction on the tree. The base case of a leaf node is trivial. For an internal node \(C_i\), assume inductively that each child set \(S^{\mathrm{top}M}(C')\) contains the top-\(M\) scores for this child's subtree. Any local solution with a score outside \(S^{\mathrm{top}M}(C')\) is dominated by at least \(M\) solutions obtained by replacing it with larger scores from \(S^{\mathrm{top}M}(C')\). Hence all top-\(M\) solutions at \(C_i\) are generated entirely from the truncated child sets. Applying the argument recursively up to the root \(C_0\) proves the claim.

\subsection{Incorporating $k_{\min}$ and $k_{\max}$ constraints}

We now extend the dynamic program to enforce explicit constraints on the number of clusters selected in the final solution. Given lower and upper limits $k_{\min}$ and $k_{\max}$, we seek clusterings $z_{C_0} \in \mathcal{Z}(C_0)$ with corresponding scores $s_{C_0} \in S(C_0)$ such that
\begin{equation}
k_{\min} \le |z_{C_0}| \le k_{\max}.
\label{eq:cluster_count_constraint}
\end{equation}

In the \emph{unconstrained} setting, optimality decomposes across subtrees: retaining only the highest-scoring clusterings in each list $S^{topM}(C_i)$ suffices to recover the globally optimal (or top-$M$) clusterings at the root. However, once constraints on the number of clusters are imposed, such as in \eqref{eq:cluster_count_constraint}, this property no longer holds. A partial clustering that is optimal within a subtree may become infeasible when combined with selections outside that subtree, while a lower-scoring partial clustering may ultimately participate in a feasible top-scoring global solution. As such, we can no longer restrict attention solely to the locally best 
partial clusterings at each node.

Instead, we retain only those partial clusterings that can still be extended to a feasible solution at the root satisfying~\eqref{eq:cluster_count_constraint}. To achieve this, we incorporate the cluster-count constraints directly into the dynamic program by computing bounds on the maximum and minimum number of clusters 
that can be selected outside each subtree without violating the fundamental coverage and non-overlap properties of flat partitions.
These bounds allow us to identify upfront, and promptly discard, clusterings that cannot participate in any feasible global solution satisfying~\eqref{eq:cluster_count_constraint}, while retaining only those that remain potentially feasible at each stage of processing. 

\medskip

\noindent
\subsubsection{Lower bound} \label{subsubsec:lowerbound}

We define a lower bound $LB_{C_i}$ representing the minimum number of clusters that \emph{must} be selected outside the subtree rooted at $C_i$ assuming that either $C_i$ or some of its sub-clusters (if any) will be selected as part of the solution.
%
This bound follows from the hierarchical structure of the tree. Any complete clustering must cover all leaves, either by selecting a leaf directly or by selecting an ancestor cluster containing it.

For a node $C_i$, the minimum valid completion outside its subtree is obtained by selecting one cluster from each sibling subtree encountered along the path from $C_i$ to the root. This follows from the fact that any partial clustering selected within $C_i$'s subtree: (a) eliminates every ancestor of $C_i$ as a candidate (non-overlap), and (b) forces at least one cluster to be selected from each subtree rooted at $C_i$'s ancestors' siblings (coverage), noting that the minimum number of selected clusters within a subtree is achieved by selecting the subtree root itself.

As a consequence, the lower bound at any $C_i$ but the root, $LB_{C_i}$ ($i \neq 0$), can be defined recursively as a function of the lower bound at $C_i$'s parent cluster. Specifically, let $\mathrm{parent}(C_i)$ denote the parent of node $C_i$, and let $\nu_{C_i}$ denote the number of sibling subtrees of $C_i$ (i.e., subtrees rooted at other children of $\mathrm{parent}(C_i)$). Each such subtree contributes at least one required cluster to any valid completion of a partial clustering selected within $C_i$'s subtree, and the same property holds true with respect to $\mathrm{parent}(C_i)$.
The lower bound can therefore be computed with a single top-down traversal of the tree. At the root,
\begin{equation}
LB_{C_0} = 0,
\end{equation}
since there are no nodes outside its subtree. Then, for a node $C_i$ in the tree,
\begin{equation}
LB_{C_i}
=
LB_{\mathrm{parent}(C_i)} + \nu_{C_i}.
\label{eq:LB_propagation}
\end{equation}

As an example, consider the tree in Figure~\ref{fig:tree_counterexample}. The lower bound for cluster 3 is $LB_{C_3} = 2$, corresponding to clusters 4 and 2 as the smallest selection ensuring complete non-overlapping coverage in a partition including cluster 3. Following an analogous reasoning, $LB_{C_2} = 1$, corresponding to cluster 1 as the smallest selection ensuring that a partition including either cluster 2 or its sub-clusters is valid.

\subsubsection{Upper bound} \label{subsubsec:upperbound}

We next define an upper bound $UB_{C_i}$ on the maximum number of clusters that can still be selected outside the subtree rooted at $C_i$, assuming that either $C_i$ or some of its sub-clusters (if any) will be selected as part of the solution.

This bound again follows from the hierarchical structure of the tree as well as the requirements of complete coverage and non-overlap of valid flat clusterings. They imply that exactly one cluster must be selected along every root-to-leaf path.
The maximum number of clusters that may still be selected outside the subtree rooted at $C_i$ is therefore equal to the number of leaves lying outside that subtree, i.e.,
\begin{equation}
UB_{C_i}
=
\mathrm{leaf_{no}}(C_0)
-
\mathrm{leaf_{no}}(C_i).
\label{eq:UB_definition}
\end{equation}

\noindent where $\mathrm{leaf_{no}}(C_i)$ is the number of leaves within the subtree rooted at $C_i$. These values can be computed with a single bottom-up traversal of the tree. If $C_i$ is a leaf node, then
\begin{equation}
\mathrm{leaf_{no}}(C_i)=1,
\end{equation}
otherwise
\begin{equation}
\mathrm{leaf_{no}}(C_i)
=
\sum_{C' \in \mathrm{ch}(C_i)}
\mathrm{leaf_{no}}(C'),
\end{equation}
where the sum is taken over the children of $C_i$, $\mathrm{ch}(C_i)$.

Considering again the example in Figure~\ref{fig:tree_counterexample}, the upper bound for cluster 3 is $UB_{C_3} = 3$, corresponding to the maximal additional selection of clusters 4, 5, and 6, whereas $UB_{C_2} = 2$, corresponding to the maximal additional selection of clusters 3 and 4.

\subsection{Feasible partial clusterings}

Given $LB_{C_i}$ and $UB_{C_i}$ defined in~\eqref{eq:LB_propagation} and~\eqref{eq:UB_definition}, we determine whether a partial clustering $z_{C_i} \in \mathcal{Z}(C_i)$ can be extended to a feasible solution at the root satisfying the cluster-count constraints in~\eqref{eq:cluster_count_constraint}.
A partial clustering can be extended to a feasible global solution if and only if
\begin{equation}
|z_{C_i}| + LB_{C_i} \le k_{\max}
\quad \text{and} \quad
|z_{C_i}| + UB_{C_i} \ge k_{\min}.
\label{eq:partial_feasibility}
\end{equation}

The first condition ensures that, even after adding the minimum number of clusters outside the subtree, the upper bound $k_{\max}$ is not exceeded. The second ensures that, even after adding the maximum possible number of clusters outside the subtree, it remains possible to reach at least $k_{\min}$. In a bottom-up processing of the tree, only partial candidate clusterings satisfying~\eqref{eq:partial_feasibility} are retained and propagated upward.

Importantly, once cluster-count constraints are imposed, partial clusterings cannot be discarded solely on the basis of their quality value. A partial clustering that is locally optimal within a subtree may be infeasible when combined with selections outside that subtree, while a lower-scoring partial clustering may yield a feasible top-scoring global solution by allowing a more favourable allocation of clusters elsewhere in the tree.

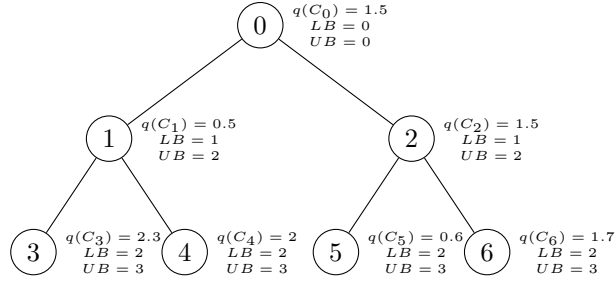
\begin{figure}
    \centering
    \begin{tikzpicture}[
          level distance=1.5cm,
          every node/.style={circle, draw, minimum size=6mm, inner sep=1pt},
          level 1/.style={sibling distance=4cm},
          level 2/.style={sibling distance=2cm}
        ]
        
        \node (C_0) {0}
          child {node (C_1) {1}
            child {node (C_3) {3}}
            child {node (C_4) {4}}
          }
          child {node (C_2) {2}
            child {node (C_5) {5}}
            child {node (C_6) {6}}
          };
        
        \clusterinfo{C_0}{0}{0}{1.5}
        \clusterinfo{C_1}{1}{2}{0.5}
        \clusterinfo{C_2}{1}{2}{1.5}
        \clusterinfo{C_3}{2}{3}{2.3}
        \clusterinfo{C_4}{2}{3}{2}
        \clusterinfo{C_5}{2}{3}{0.6}
        \clusterinfo{C_6}{2}{3}{1.7}
        
    \end{tikzpicture} 
    \caption{Illustrative example of a cluster tree with $N$=7 nodes where the quality, lower bound, and upper bound for each node is displayed.} 
    \label{fig:tree_counterexample}
\end{figure}

This behaviour can be illustrated using the example in Figure~\ref{fig:tree_counterexample}, with constraints on the number of clusters given by $k_{\min}=2$ and $k_{\max}=3$. We can show that, under these constraints, pruning solely on the basis of local quality value may eliminate partial clusterings required to construct the globally optimal feasible solution.

Processing the tree bottom--up, locally at each leaf node the only possible partial candidate clustering is the leaf cluster itself. For instance, for cluster 3 the only candidate clustering is $z_{C_3} = \{C_3\}$, with $|z_{C_3}| = 1$. 
Since $LB_{C_3} = 2$ and $UB_{C_3} = 3$, then both conditions in \eqref{eq:partial_feasibility} are satisfied and this partial local solution cannot be eliminated.
By construction, this must be the same for $C_3$'s sibling, $C_4$, which is also a leaf. Applying the test to the other leaves, $C_5$ and $C_6$, will see both also clearing \eqref{eq:partial_feasibility}. 

We can then move upward to the internal, ancestor nodes. Starting at node $C_1$, there are two partial candidate clusterings $z_{C_1}$ in $\mathcal{Z}(C_1)$:
\[
\{C_1\}, \qquad q(\{C_1\}) = 0.5,
\]
with cardinality $|z_{C_1}| = |\{C_1\}| = 1$, and
\[
\{C_3, C_4\}, \qquad q(\{C_3, C_4\}) = 4.3.
\]
with cardinality $|z_{C_1}| = |\{C_3,C_4\}| = 2$. Both can be extended to feasible global solutions as they satisfy \eqref{eq:partial_feasibility} given $LB_{C_1} = 1$ and $UB_{C_1} = 2$, and thus neither can be locally pruned based on the cluster-count constraints. Under a local top-$M$ selection with $M=1$ (as implemented in the unconstrained top-$M$ strategy), however, only $\{C_3, C_4\}$ would be retained as it has a higher quality score.

At node $C_2$, we similarly obtain:
\[
\{C_2\}, \qquad q(\{C_2\}) = 1.5,
\]
and
\[
\{C_5, C_6\}, \qquad q(\{C_5, C_6\}) = 2.3.
\]
Again, both are feasible at this stage, but a local top-1 selection as implemented in the unconstrained approach would only retain $\{C_5, C_6\}$, locally discarding the lower-scoring alternative, $\{C_2\}$.

At the root $C_0$, cross-combining all feasible partial clusterings from the left and right subtrees yields the following candidate global clusterings $z_{C_0}$ in $\mathcal{Z}(C_0)$:
\[
\begin{aligned}
\{C_1, C_2\} &:\; q = 2.0, \\
\{C_1, C_5, C_6\} &:\; q = 2.8, \\
\{C_2, C_3, C_4\} &:\; q = 5.8, \\
\{C_3, C_4, C_5, C_6\} &:\; q = 6.6.
\end{aligned}
\]

The highest-scoring solution $\{C_3, C_4, C_5, C_6\}$ violates $k_{\max}=3$ and is therefore infeasible. The best feasible solution is $\{C_2, C_3, C_4\}$, with score $5.8$. However, this solution requires the partial clustering $\{C_2\}$ from node $C_2$, which would have been discarded should pruning have been locally applied based on top quality scores only.

Thus, although $\{C_5, C_6\}$ is locally optimal at node $C_2$, eliminating $\{C_2\}$ would remove the globally optimal feasible solution. \emph{This illustrates that, under $[k_{\min}$,$k_{\max}]$ constraints, feasibility depends on combinations across subtrees, and the unconstrained top-$M$ local pruning strategy based solely on quality scores is no longer valid.}

\subsection{Efficient combination of top-$M$ and $[k_{\min},k_{\max}]$ constraints} \label{subsec:efficient_combination}

To efficiently compute the top-$M$ solutions under cluster-count constraints, we must identify partial clusterings that cannot participate in any globally feasible top-$M$ solution at the root $C_0$, and discard them during the dynamic 
program.
Without such pruning, all feasible partial clusterings at each node would need to be retained, leading to a combinatorial number of candidates. We therefore introduce a dominance-based pruning strategy that compares partial clusterings not only by quality score but also by the number of additional clusters they permit outside their subtree.

We first consider the case with $M=1$. Let $z_{C_i}^{(a)}, z_{C_i}^{(b)} \in \mathcal{Z}(C_i)$ be two partial clusterings with corresponding scores $s_{C_i}^{(a)}, s_{C_i}^{(b)} \in S(C_i)$. Suppose there exists some number of additional clusters $k$, such that both partial clusterings can be extended to feasible global solutions using the same number $k$ of additional clusters, i.e.,
\begin{equation}
k_{\min} \le |z_{C_i}^{(a)}| + k \le k_{\max},
\quad
k_{\min} \le |z_{C_i}^{(b)}| + k \le k_{\max}.
\label{eq:dominance_feasible_completion}
\end{equation}

If \( s_{C_i}^{(a)} > s_{C_i}^{(b)} \), then any feasible completion of $z_{C_i}^{(b)}$ using $k$ additional clusters can be matched by a completion of $z_{C_i}^{(a)}$ with strictly higher quality score. Hence, \emph{for this value of} $k$, $z_{C_i}^{(b)}$ cannot yield an optimal feasible solution, as it is dominated by $z_{C_i}^{(a)}$.

A trivial case arises when the two partial clusterings have the same size, i.e., $|z_{C_i}^{(a)}| = |z_{C_i}^{(b)}|$. In this case, both admit exactly the same feasible completions (\emph{for any admissible} $k$), and therefore, if $s_{C_i}^{(a)} > s_{C_i}^{(b)}$, the clustering $z_{C_i}^{(b)}$ can be safely discarded.

We can further generalise this by determining, for each partial clustering $z_{C_i} \in \mathcal{Z}(C_i)$, the \emph{range for the number of additional clusters} that may be selected outside the subtree rooted at $C_i$ while preserving feasibility.

The minimum admissible number of additional clusters is constrained both by the structural lower bound ensuring valid flat solutions as well as 
the user-defined lower bound delineating feasible solutions. Specifically, we must select: (a) at least $LB_{C_i}$ counted clusters outside the subtree rooted at $C_i$ to ensure coverage; AND (b) 
at least $k_{\min} - |z_{C_i}|$ counted clusters in order to satisfy the global constraint \eqref{eq:cluster_count_constraint}. Thus, the minimum admissible number of additional clusters is
\begin{equation}
\ell\!\left(z_{C_i}\right)
=
\max\bigl(
LB_{C_i},\;
k_{\min} - |z_{C_i}|
\bigr).
\label{eq:ell_definition}
\end{equation}

Similarly, the maximum admissible number of additional clusters is constrained both by the structural upper bound and by the requirement not to exceed $k_{\max}$. 
We cannot select more than $UB_{C_i}$ counted clusters outside the subtree rooted at $C_i$. 
In addition, we must ensure that the total number of counted clusters does not exceed $k_{\max}$, that is, 
we can select at most $k_{\max} - |z_{C_i}|$ counted clusters from outside $C_i$'s subtree.
Therefore, the maximum admissible number of additional clusters is
\begin{equation}
u\!\left(z_{C_i}\right)
=
\min\bigl(
UB_{C_i},\;
k_{\max} - |z_{C_i}|
\bigr).
\label{eq:u_definition}
\end{equation}

Thus, the range of additional clusters that may be added to the partial clustering $z_{C_i}$ in order to form a feasible global solution is given by
\begin{equation}
\bigl[
\ell\!\left(z_{C_i}\right),\,
u\!\left(z_{C_i}\right)
\bigr].
\label{eq:completion_interval}
\end{equation}


At node $C_i$, the elements of $\mathcal{Z}(C_i)$ are processed in decreasing order of quality score. A partial clustering $z_{C_i} \in \mathcal{Z}(C_i)$ is retained only if: (i) its feasible completion interval~\eqref{eq:completion_interval} is not empty, i.e., $\ell\!\left(z_{C_i}\right) \leq
u\!\left(z_{C_i}\right)$; \emph{and} (ii) there is some portion of this interval that is not already covered by a higher-scoring retained solution. If the latter is not satisfied, then $z_{C_i}$ can be safely discarded because, for any admissible number of additional clusters $k$ outside $C_i$'s subtree, there exists at least one higher-scoring $z'_{C_i} \in \mathcal{Z}(C_i)$ that will admit the same feasible completion, which by construction will necessarily have a higher final global score. Formally, $z_{C_i}$ is retained only if \mbox{$\ell\!\left(z_{C_i}\right) \leq
u\!\left(z_{C_i}\right)$} and
\begin{equation}
[\ell(z_{C_i}), u(z_{C_i})]
\not\subseteq
\bigcup_{z'_{C_i} \succ z_{C_i}}
[\ell(z'_{C_i}), u(z'_{C_i})],
\end{equation}
where $z'_{C_i} \succ z_{C_i}$ denotes a higher-scoring partial clustering at node $C_i$. Otherwise, $z_{C_i}$ is safely discarded either as infeasible or as a dominated candidate.

For the general top-$M$ setting, we retain up to $M$ highest-scoring partial clusterings \emph{for each admissible completion size}. Once again, for the sake of notational simplicity, although clusterings are the underlying objects of interest, the formulation can be expressed in terms of their quality scores. Specifically, given a set of scores of partial candidate clusterings with associated feasible completion intervals~\eqref{eq:completion_interval}, we define $\mathrm{TopM}^{\mathrm{feas}}(\cdot)$ as the corresponding dominance-based selection operator. $\mathrm{TopM}^{\mathrm{feas}}$ processes local candidates in decreasing order of score, discarding: (i) any infeasible candidate $z_{C_i}$, whenever \mbox{$\ell\!\left(z_{C_i}\right) >
u\!\left(z_{C_i}\right)$}; and (ii) any dominated candidate $z_{C_i}$, whenever all admissible completion sizes within its feasible completion interval (i.e., every integer $k$ within \eqref{eq:completion_interval}) is fully covered by $M$ or more higher-scoring retained candidates.

Let $S^{Mfeas}(C_i)$ denote the set of scores retained after applying $\mathrm{TopM}^{\mathrm{feas}}$ at node $C_i$. 
Replacing $S^{topM}(C_i)$ with $S^{Mfeas}(C_i)$ and $\mathrm{TopM}$ with $\mathrm{TopM}^{\mathrm{feas}}$ in Equations~\eqref{eq:sum} and \eqref{eq:recursion} yields
\begin{equation}
\bigoplus_{C' \in \mathrm{ch}(C_i)} S^{Mfeas}(C')
=
\left\{
\sum_{C' \in \mathrm{ch}(C_i)} s_{C'}
\;\middle|\;
s_{C'} \in S^{Mfeas}(C')
\right\}.
\end{equation}

\noindent and

\begin{equation}
S^{Mfeas}(C_i)
=
\mathrm{TopM}^{\mathrm{feas}}\!\left(
\{ q(C_i) \}
\;\cup\;
\bigoplus_{C' \in \mathrm{ch}(C_i)} S^{Mfeas}(C')
\right).
\end{equation}

The solution at the root, $S^{Mfeas}(C_0)$, yields the top-$M$ feasible scores associated with clusterings satisfying the $[k_{\min}$,$k_{\max}]$ constraints.

\subsection{Pruning example}

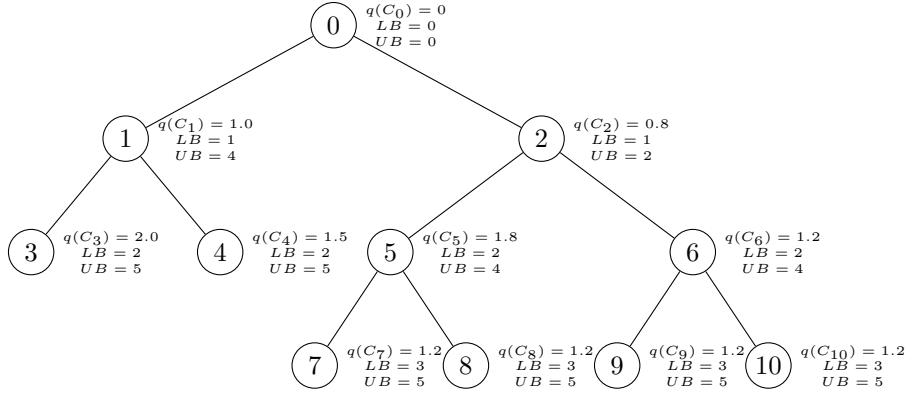
\begin{figure}
    \centering
    \begin{tikzpicture}[
      level distance=1.5cm,
      every node/.style={circle, draw, minimum size=6mm, inner sep=1pt},
      level 1/.style={sibling distance=5.5cm},
      level 2/.style={sibling distance=2.5cm},
      level 3/.style={sibling distance=2.2cm}
    ]
    
    \node (C_0) {0}
      child {node (C_1) {1}
        child {node (C_3) {3}}
        child {node (C_4) {4}}
      }
      child {node (C_2) {2}
        child[sibling distance=4cm] {node (C_5) {5}
            child[sibling distance=2cm] {node (C_7) {7}}
            child[sibling distance=2cm] {node (C_8) {8}}}
        child[sibling distance=4cm] {node (C_6) {6}
            child[sibling distance=2cm] {node (C_9) {9}}
            child[sibling distance=2cm] {node (C_{10}) {10}}}
      };
    
    \clusterinfo{C_0}{0}{0}{0}
    \clusterinfo{C_1}{1}{4}{1.0}
    \clusterinfo{C_2}{1}{2}{0.8}
    \clusterinfo{C_3}{2}{5}{2.0}
    \clusterinfo{C_4}{2}{5}{1.5}
    \clusterinfo{C_5}{2}{4}{1.8}
    \clusterinfo{C_6}{2}{4}{1.2}
    \clusterinfo{C_7}{3}{5}{1.2}
    \clusterinfo{C_8}{3}{5}{1.2}
    \clusterinfo{C_9}{3}{5}{1.2}
    \clusterinfo{C_{10}}{3}{5}{1.2}
    
    \end{tikzpicture}
    \caption{Illustrative example of a cluster tree with $N=11$ nodes where the quality, lower bound, and upper bound for each node is displayed.}
    \label{fig:tree_kbounds}
\end{figure}

To illustrate the dominance-pruning procedure, consider the example tree shown in Figure~\ref{fig:tree_kbounds} with $M=1$, $k_{\min}=3$, and $k_{\max}=4$. Table~\ref{tab:dominance_example} shows the candidate partial clusterings generated at node $C_2$ and the pruning decisions applied to them. For this node, the structural bounds $LB_{C_2}$ and $UB_{C_2}$ imply that any feasible completion must select either one or two additional clusters outside the subtree rooted at $C_2$.

The candidate partial clusterings are processed in decreasing order of quality. For each partial clustering, the feasible interval \([\ell,u]\) is computed using Equations~\eqref{eq:ell_definition} and~\eqref{eq:u_definition}, representing the admissible numbers of additional clusters that may be selected outside the subtree while still satisfying the global cluster-count constraints. 

Starting with the highest-scoring clustering (row 1 of Table~\ref{tab:dominance_example}, $\{C_7,C_8,C_9,C_{10}\}$), we compute $1 = \ell > u = 0$, which represents an invalid (empty) feasible interval $[\ell,u]$. Since no admissible completion exists for this interval, the clustering is infeasible and, as such, it is promptly discarded.
Proceeding to the next highest-scoring clustering (row 2, $\{C_5,C_9,C_{10}\}$), we obtain a feasible interval of $[1,1]$. Since this interval contains an admissible completion size (i.e., 1) that is not yet covered by at least $M = 1$ higher-scoring retained candidates, the clustering is retained and the covered completion sizes are updated to $\{1\}$.
For the third clustering (row 3, $\{C_6,C_7,C_8\}$), the feasible interval is again $[1,1]$. However, this interval is already covered by the higher-scoring retained clustering in row 2. Consequently, the clustering is dominated and discarded.
Considering the fourth clustering (row 4, $\{C_5,C_6\}$), we obtain a feasible interval of $[1,2]$. Since completion size $2$ is not yet covered by any retained clustering, the interval is not fully covered and the clustering is retained. The covered completion sizes are therefore updated to $\{1,2\}$.
Finally, for the clustering consisting of $C_2$ itself (row 5), the feasible interval is $[2,2]$. This interval is already covered by previously retained clusterings, and therefore the clustering is dominated and discarded. The remaining retained clusterings ($\{C_5,C_9,C_{10}\}$ and $\{C_5,C_6\}$) correspond to the highest-scoring solutions required to preserve top-$M$ constrained global optimality upward the tree.

In this example, the number of retained partial clusterings is reduced from five to two at node $C_2$. 
This dominance pruning is what enables efficient computation of the top-$M$ solutions under $[k_{\min}$,$k_{\max}]$ constraints by substantially reducing the number of partial candidate clusterings retained at each node.

\begin{table}[h] 
\centering
\caption{Example of dominance pruning at node $C_2$ of the tree in Figure~\ref{fig:tree_kbounds}, with $M=1$, $k_{\min}=3$ and $k_{\max}=4$. Candidate partial clusterings are ranked in decreasing order of score. The feasible range $[\ell,u]$ gives the admissible numbers of additional clusters that may be selected outside the subtree. The ``Covered values of $k$'' column records completion sizes already covered by at least $M = 1$ higher-scoring retained candidates. A candidate is retained only if its feasible range is not fully covered; otherwise it is dominated.}
\label{tab:dominance_example}
\begin{tabular}{c c c c c l}
\hline
Rank & \makecell{Partial clustering \\ at $C_2$} & Score & \makecell{Feasible range\\ $[\ell,u]$} & \makecell{Covered values \\ of $k$} & Status \\
\hline
1 & $\{C_7,C_8,C_9,C_{10}\}$ & 4.8 & $[-,-]$ & \{\,\} & Infeasible \\
2 & $\{C_5,C_9,C_{10}\}$     & 4.2 & $[1,1]$ & \{1\} &  Retained \\
3 & $\{C_6,C_7,C_8\}$        & 3.6 & $[1,1]$ & \{1\} &  Dominated by Rank 2 \\
4 & $\{C_5,C_6\}$            & 3.0 & $[1,2]$ & \{1,2\} &  Retained \\
5 & $\{C_2\}$                & 0.8 & $[2,2]$ & \{1,2\} &  Dominated by Ranks 2 and 4 \\
\hline
\end{tabular}
\end{table}

\subsection{Overall algorithmic procedure} 

The preceding sections describe the individual components of the FOSC-X framework. We now summarise how these components are combined into the complete flat clustering extraction procedure, shown in Algorithm~\ref{alg:FOSC-X}.

The algorithm assumes as input a hierarchical cluster tree $\mathcal{T}$ together with a quality value for each node, $q(C_i)$, which needs to be precomputed according to a suitable cluster quality measure $Q$ satisfying the properties of locality and additivity previously discussed (Section~\ref{sec:backquality}). Prior to flat clustering extraction, the structural bounds $LB_{C_i}$ and $UB_{C_i}$ must also be precomputed for each node based on two traversals of the tree (Sections~\ref{subsubsec:lowerbound} and \ref{subsubsec:upperbound}), providing all the information required as input to the subsequent dynamic programming procedure in Algorithm~\ref{alg:FOSC-X}.

Algorithm~\ref{alg:FOSC-X} performs a bottom-up traversal of the cluster tree. At each node $C_i$, a set $\mathcal{Z}(C_i)$ of candidate partial clusterings is maintained. Leaf nodes initialise this set with the partial clustering consisting only of the node itself. For each internal node $C_i$, augmented partial candidate clusterings at $C_i$ are generated by combining partial clusterings propagated from the children of $C_i$, in addition to the alternative of selecting $\{C_i\}$ itself instead.
The resulting candidates are then processed according to the cluster-count constraints and dominance-pruning procedure described in Section~\ref{subsec:efficient_combination}, as summarised in Algorithm~\ref{alg:dominance}. Infeasible and dominated candidates are discarded --- only feasible, non-dominated candidates strictly required to guarantee preservation of the final (global) top-$M$ feasible solutions are locally retained. Back in Algorithm~\ref{alg:FOSC-X}, these surviving candidates are subsequently propagated to the parent node.

Once the root node has been processed, the set $\mathcal{Z}(C_0)$ contains the top-$M$ feasible global flat clusterings, which are returned as the final solution(s).

\begin{algorithm}
\caption{FOSC-X}
\label{alg:FOSC-X}
\begin{algorithmic}[1]
\Require Cluster tree $\mathcal{T}$, cluster quality values $\{q(C_i): C_i \in \mathcal{T}\}$, no. of top solutions $M$, (optional) cluster-count constraints $k_{\min}, k_{\max}$ and node bounds $LB_{C_i}, UB_{C_i}$

\ForAll{nodes $C_i$ in postorder traversal of $\mathcal{T}$}

    \If{$C_i$ is a leaf}
        \State $Candidates[C_i] \gets \{\text{clustering containing only } C_i\}$
        \State \textbf{continue}
    \EndIf

    \State Generate all combinations formed by selecting one candidate from the list of candidates $Candidates[\mathrm{ch}(C_i)]$ of each child of $C_i$, $\mathrm{ch}(C_i)$
    \State Compute the quality score for each such a combination as the sum of the individual quality scores of its merged constituents
    \State Include all these combinations into $Candidates[C_i]$
    \State Add the alternative candidate clustering containing only $C_i$ to $Candidates[C_i]$

    \State Sort $Candidates[C_i]$ by decreasing quality score

    \If{cluster-count constraints are specified}
        \State $Candidates[C_i] \gets \textsc{DominancePrune}(Candidates[C_i])$
    \Else
        \State $Candidates[C_i] \gets$ the top-$M$ highest-scoring candidates in $Candidates[C_i]$
    \EndIf

\EndFor

\State \Return $Candidates[\mathrm{root}]$

\end{algorithmic}
\end{algorithm}

\begin{algorithm}
\caption{DominancePrune}
\label{alg:dominance}
\begin{algorithmic}[1]

\Require Candidate set $Candidates[C_i]$ sorted by decreasing quality score, no. of top solutions $M$, cluster-count constraints $k_{\min}, k_{\max}$, node bounds $LB_{C_i}, UB_{C_i}$

\State $TopCandidates \gets \emptyset$

\State Initialise $coverage[k]$ to zero for all completion sizes $k \in [LB_{C_i},\,UB_{C_i}]$

\ForAll{candidate clusterings $z_{C_i}$ in $Candidates[C_i]$}

    \State Compute the feasible completion interval $[\ell(z_{C_i}),u(z_{C_i})]$ in Eq.~\eqref{eq:completion_interval}

    \If{$\ell(z_{C_i}) \leq u(z_{C_i})$ AND there exists an integer
    $k \in [\ell(z_{C_i}),u(z_{C_i})]$
    such that $coverage[k] < M$} 

        \State Add $z_{C_i}$ to $TopCandidates$

        \ForAll{$k \in [\ell(z_{C_i}),u(z_{C_i})]$}
            \State $coverage[k] \gets coverage[k] + 1$
        \EndFor

    \EndIf

\EndFor

\State \Return $TopCandidates$

\end{algorithmic}
\end{algorithm}

\section{Noise and Tree Condensation} \label{sec:noise}

\subsection{Noise} \label{subsec:noise}

In flat clustering, noise refers to observations that are not considered to belong meaningfully to any cluster. Rather than forming part of the clustering structure, such observations are intentionally excluded from all selected clusters and therefore do not contribute to the cluster count. Noise thus provides a mechanism for representing observations that do not strongly belong to any cluster.

Noise may also occur in hierarchical clustering, possibly arising through several mechanisms. Density-based clustering methods such as HDBSCAN* \cite{campello2013density,campello2015hierarchical} naturally identify as noise those observations that no longer belong to any sufficiently dense cluster according to a varying density threshold corresponding to the different hierarchical levels. Tree condensation procedures \cite{campello2013fosc} may in turn designate cluster nodes below a minimum size threshold as ``not-a-cluster'', case in which all the observations within the corresponding cluster subtree are deemed as noise. Noise may also arise through preprocessing steps such as outlier detection or by arbitrarily designating singleton observations as noise rather than as valid singleton clusters.


\emph{Regardless of its source, FOSC-X treats noise uniformly through its representation within the hierarchy, as singleton leaf observations designated as noise in the cluster tree $\mathcal{T}$ provided as input}. These noise leaves do not contribute to the number of selected clusters in a flat clustering (although they may still contribute to the objective function used for extraction, depending on the chosen quality measure). As a consequence, noise affects computation of the feasibility bounds in the constrained extraction scenario.

Noise affects the structural lower bound $LB_{C_i}$ on the minimum number of counted clusters required outside the subtree rooted at $C_i$. In the noiseless case, every root-to-leaf path must contribute exactly one counted cluster to any feasible solution. When noise leaves are present, however, selecting a noise leaf remains valid while contributing zero to the cluster count. Consequently, subtrees for which all leaves consist of noise no longer require the selection of counted clusters and are excluded from the sibling count $\nu_{C_i}$ in Equation~\eqref{eq:LB_propagation}. Noise therefore relaxes the structural requirement that every subtree contributes at least one counted cluster (ensuring local coverage), which enlarges the feasible solution space under $k_{\max}$ constraints, potentially permitting higher-scoring partial clusterings that would otherwise be infeasible.


Noise also affects the structural upper bound $UB_{C_i}$ on the maximum number of counted clusters allowed outside the subtree rooted at $C_i$. In the noiseless setting, this is obtained by selecting all leaves of the hierarchy outside the subtree in question. When noise is present, noise leaves do not contribute to the cluster count, so the structural upper bound $UB_{C_i}$ is computed as the number of \emph{non-noise} leaves of the hierarchy outside the subtree rooted at $C_i$, i.e., the number of nodes outside the subtree in question that represent terminal nodes after conceptually trimming off all the noise. These correspond to the deepest non-noise cluster nodes outside $C_i$'s subtree.

Importantly, the dynamic programming recursion and dominance pruning strategy of FOSC-X remain unchanged; noise affects the feasibility conditions used during propagation through a modified cluster counting policy for candidate solutions and, accordingly, for structural bounds, without affecting the extraction algorithm itself.

\subsection{Tree condensation} \label{sec:condense}

Tree condensation is a common preprocessing step in density-based hierarchical clustering. In HDBSCAN*, condensation can be performed in two different ways. First, by not considering as a cluster split (but as cluster shrinkage instead) when data observations simply detach from a cluster as they become noise according to an increasing minimum density threshold. Second, by possibly enforcing a minimum cluster size constraint. In the latter case, when a cluster splits, any child cluster containing fewer than the required minimum number of observations is not retained as a valid cluster in the hierarchy. Instead, the branch corresponding to that child terminates, and its observations are treated as noise from that point onward in the hierarchy. This procedure was introduced in HDBSCAN* \cite{campello2013density} and generalised to arbitrary hierarchical trees, including traditional dendrograms, in the original FOSC framework \cite{campello2013fosc}.

The effect of condensation is to simplify the hierarchy by removing small or insignificant nodes, thereby reducing the number of candidate clusters considered during flat cluster extraction and reducing the search space explored by FOSC/FOSC-X. Condensation introduces noise into the hierarchy through the termination of condensed branches, which is handled as described in Section~\ref{subsec:noise}.

Condensation may also alter the values assigned by some quality measures. In particular, Stability and its density-based counterpart Excess of Mass are defined with respect to the distance/density levels at which clusters appear and disappear within the hierarchy. Condensation modifies these cluster lifetimes and therefore changes the resulting objective values assigned to candidate clusters (for details, please see \cite{campello2013fosc, campello2015hierarchical}). 

\section{Computational complexity} \label{subsec:complexity}

The dominant computational cost of FOSC-X arises from generating and sorting the partial candidate clusterings constructed at each node. A detailed derivation of the complexity is provided in Appendix~\ref{sec:complexity}; here we present an intuitive overview.

\subsection{Upper bound on the number of retained partial clusterings} \label{sec:upper_bound}

The dominance pruning procedure induces an upper bound on the number of partial candidate clusterings retained at any node $C_i$.
As established previously, partial candidate clusterings containing the same number of clusters admit identical feasible completions, implying that for each admissible cluster count only the $M$ highest-scoring such partial clusterings can participate in an optimal global solution, while all others are dominated and discarded. Since any feasible global solution contains at most $k_{\max}$ clusters, a partial candidate clustering within a subtree may also contain at most $k_{\max}$ clusters. Consequently, there exist at most $k_{\max}$ possible subtree cardinalities, yielding the bound \( |\mathcal{Z}(C_i)| \le M k_{\max}. \)
This bound is tight in the worst case, where each retained partial candidate clustering contains a distinct number of clusters.

\subsection{Complexity analysis}

In the unconstrained top-$M$ setting, each child propagates upwards at most $M$ partial candidate clusterings. For a binary cluster tree (or more generally, a tree with a finite branching factor, since any non-binary tree with a finite branching factor can be represented as a binary tree that will produce the same flat extraction), combining child solutions yields at most $M^2$ candidates at each internal node. Sorting these candidates requires $O(M^2 \log M)$ time. Since the dynamic program processes each node exactly once, then the total complexity is $O(N M^2 \log M)$, where $N$ is the number of nodes in the tree; and since trees corresponding to clustering hierarchies constructed over $n$ observations contain $O(n)$ nodes, this translates to $O(n M^2 \log M)$. 

Condensation preserves this asymptotic bound, since it only removes nodes from the hierarchy and therefore cannot increase the tree size, $N$. Although condensation may substantially reduce the number of nodes relative to the dataset size in practice, the complexity remains bounded by $O(n)$ with respect to the number of observations.

Under cluster-count constraints, dominance pruning ensures that at most $Mk_{\max}$ partial candidate clusterings are retained at any node. In the worst case, combining child solutions may therefore generate up to $(Mk_{\max})^2$ candidate combinations at a node. However, this worst case cannot occur uniformly across the tree. A node can only propagate $Mk_{\max}$ candidates if its subtree is sufficiently large to support $k_{\max}$ counted clusters, and the number of such nodes within a tree decreases proportionally as $k_{\max}$ increases. Although the per-node worst-case cost is quadratic in $Mk_{\max}$, the aggregate cost across the hierarchy scales as $O(N k_{\max} M^2 \log(k_{\max}M))$ as a function of the tree size, or $O(n k_{\max} M^2 \log(k_{\max}M))$ as a function of the dataset size. A formal derivation of this amortised bound is provided in Appendix~\ref{sec:complexity}. The constrained-case complexity is a worst-case upper bound; in practice, the number of retained partial clusterings is typically much smaller than the theoretical maximum.

When $M$ and $k_{\max}$ are treated as user-defined fixed parameters, the FOSC-X flat extraction procedure is therefore linear in the number of observations in the dataset.

\section{Experimental Evaluation} \label{sec:eval}

\subsection{Benchmark description}

In the following we will produce benchmarks for the top-$M$, constrained $[k_{\min}, k_{\max}]$, and combined (constrained top-$M$) methods for the FOSC-X framework. \footnote{Code and datasets available at \url{https://github.com/Campello-Lab/FOSC-X}} 

\subsubsection{Datasets}

We generate 2100 synthetic clustering datasets using the \texttt{scikit-learn} library \cite{scikit-learn}. Each dataset contains $n=2000$ observations, and for every configuration we generate $20$ independent datasets. The number of ground-truth clusters varies over $k^* \in \{5,10,15,20,25\}$ and the dimensionality over $dim \in \{2,5,10,15,20,50,200\}$.
For each parameter combination we consider three data types: isotropic Gaussian clusters, anisotropic clusters obtained by applying a random linear transformation, and clusters with heterogeneous variances where cluster standard deviations are sampled uniformly from $[0.1,5.0]$.

\subsubsection{Clustering algorithms}

We evaluate FOSC-X on hierarchies produced by two widely used clustering approaches, both equipped with Euclidean distance as base metric: Ward-Linkage, a classic agglomerative variance-minimisation method that produces a binary cluster tree, and HDBSCAN*, a density-based hierarchical algorithm. 

%
For Ward-Linkage, we use the implementation provided by the \texttt{scikit-learn} library \cite{scikit-learn}, and the resulting hierarchy is used directly as the input cluster tree into FOSC-X.

For HDBSCAN*, we use the Python implementation of \cite{McInnes2017} with its default setting $\texttt{min\_samples} = \texttt{min\_cluster\_size} = 5$. The condensed cluster tree returned by HDBSCAN* is used as the input cluster tree for FOSC-X.

\subsubsection{Quality measure and evaluation criteria}

In all experiments, the quality measure used as objective function for flat clustering extraction with both the original FOSC as well as FOSC-X is \emph{cluster Stability} \cite{campello2013fosc} (Section~\ref{sec:backquality}), which corresponds to the \emph{Excess of Mass (EOM)} criterion in the case of HDBSCAN*'s condensed cluster trees \cite{campello2013density, campello2015hierarchical}. It is worth noting that the goal of these experiments is not to compare different quality measures, but instead show and quantify experimentally the superiority of FOSC-X over FOSC. As the extraction framework is agnostic to the choice of cluster quality measure, we adopt cluster Stability throughout. This allows us to focus on evaluating the different extraction strategies rather than differences between alternative quality criteria.

Clustering performance with respect to ground truth labels (available for all datasets) is evaluated using Adjusted Mutual Information (AMI) \cite{JMLR:v11:vinh10a}. For methods that produce multiple clusterings (i.e., top-$M$ with $M > 1$), all returned clusterings are evaluated and the maximum AMI is reported.

For condensed cluster trees containing noise (HDBSCAN*), the AMI is initially computed only on the subset of data points assigned to clusters (i.e., excluding points labelled as noise). To account for the proportion of data clustered, the resulting AMI score is then multiplied by the fraction of data within that subset, which corresponds to penalising the remaining fraction of data left unclustered as noise.

For HDBSCAN*, hierarchy condensation may prevent a feasible solution from existing when cluster-count constraints are used. In particular, the condensed hierarchy may contain fewer cluster leaves than required by the specified lower bound $k_{\min}$, making it impossible to satisfy the constraints. In such cases, no AMI value is assigned for that dataset, and it is excluded from the statistical analyses.

Statistical significance across methods is assessed using the Friedman test followed by the Nemenyi post-hoc test \cite{Demvsar2006}. Methods are ranked on each dataset according to AMI, and the Friedman test evaluates whether the average ranks differ significantly without assuming normality. When significance is detected, the Nemenyi test performs pairwise comparisons. Results are visualised using Critical Difference diagrams, where methods whose rank differences are smaller than the critical difference are connected, indicating no statistically significant difference.

\subsubsection{Evaluation scenarios}

FOSC-X provides the user with substantially greater control over the clustering extracted from a cluster tree. Therefore, we evaluate it under several different settings designed to reflect typical use cases enabled by its flexibility, including scenarios with and without cluster-count constraints and when the number of ground-truth clusters ($k^*$) is known exactly or approximately. Each case is compared to the baseline of $M=1$ without cluster-count constraints, corresponding to the original FOSC result.

In short, FOSC-X is applied under the following configurations:

\begin{enumerate}
    \item \textbf{Baseline FOSC:}  
    $M = 1$, $k_{\min} = 2$, $k_{\max} = n$.  
    This setting is equivalent to the original FOSC method and serves as the primary baseline.

    \item \textbf{top-$M$:}  
    $M = [5,10]$, $k_{\min} = 2$, $k_{\max} = n$.  
    This evaluates the benefit of enumerating multiple globally optimal and near-optimal solutions without constraints for the number of clusters. This is tested for both $M=5$ and $M=10$ to compare the benefit of exploring different numbers of top-scoring solutions. 

    \item \textbf{Fixed $k$:}  
    $M = 1$, $k_{\min} = k^*$, $k_{\max} = k^*$,  
    where $k^*$ is the ground-truth number of clusters.  
    This isolates the effect of enforcing an exact cluster count.

    \item \textbf{Relaxed $k$:}  
    $M = 1$, $k_{\min} = k^* - 3$, $k_{\max} = k^* + 3$.  
    This evaluates robustness when the true number of clusters is only approximately known.

    \item \textbf{top-$M$ with Fixed $k$:}  
    $M = 5$, $k_{\min} = k^*$, $k_{\max} = k^*$.  
    This combines multi-solution enumeration with exact cluster-count constraints.

\end{enumerate}

The Relaxed $k$ and Fixed $k$ configurations with $M=1$ correspond to settings that could also be addressed using the DynaCut method \cite{marti2017cut}. Although the original DynaCut formulation assumes centre-based and separable quality measures, we observe that the optimisation procedure in fact only requires locality and additivity. Consequently, FOSC objectives such as Stability are also compatible with the DynaCut framework. Under this interpretation, Fixed $k$ corresponds to determining the optimal clustering for a single prescribed value of $k$, while Relaxed $k$ is equivalent to evaluating a range of candidate values and selecting the highest-scoring solution among them. Since these settings arise as special cases of the more general FOSC-X formulation, we do not perform a separate comparison against DynaCut (except in Section~\ref{sec:empiricalComp}, when they are compared in terms of runtime). 

\subsection{Results} \label{subsubsec:synthetic_Results}

\begin{figure}[!h]
	\centering
	\includegraphics[width=\textwidth, clip]{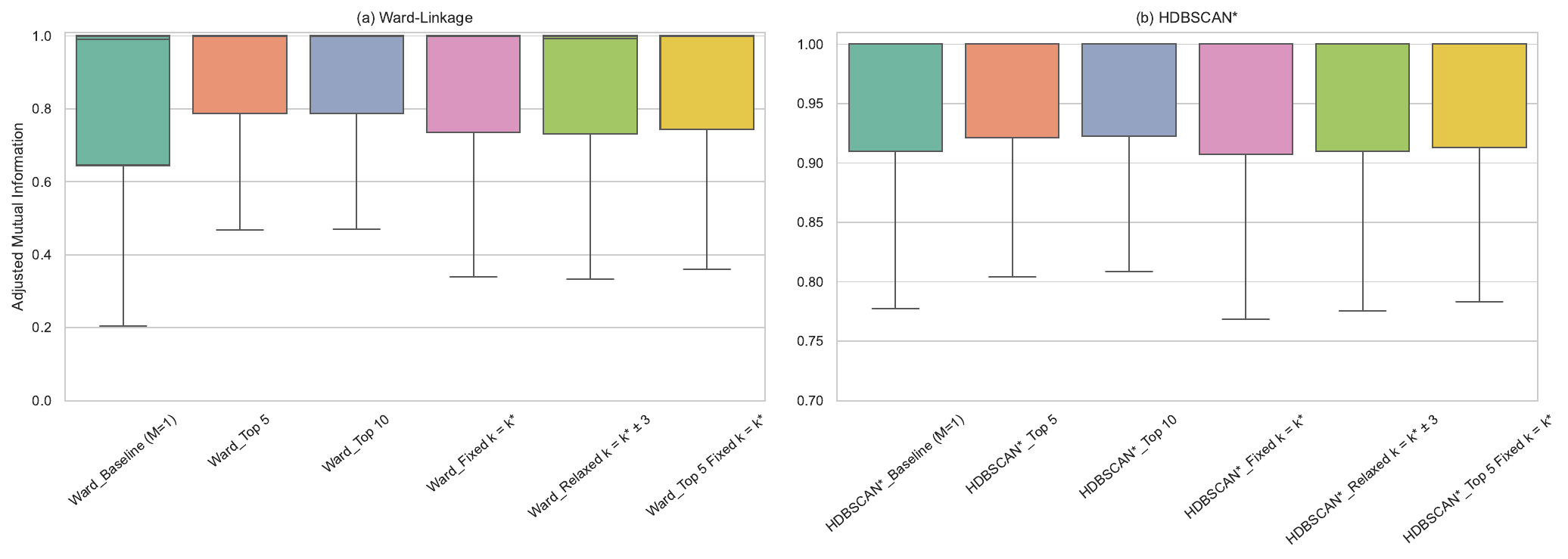}
	\caption{Distribution of Adjusted Mutual Information (AMI) scores across the six FOSC-X cluster extraction scenarios on the synthetic datasets. Results using: (a) Ward-Linkage; (b) HDBSCAN*. For interpretability, the y-axis in panel (b) is scaled within the range $[0.7, 1.0]$.}
	\label{fig:box_synthetic}
\end{figure}

\begin{figure}[!h]
	\centering
	\includegraphics[width=\textwidth, clip]{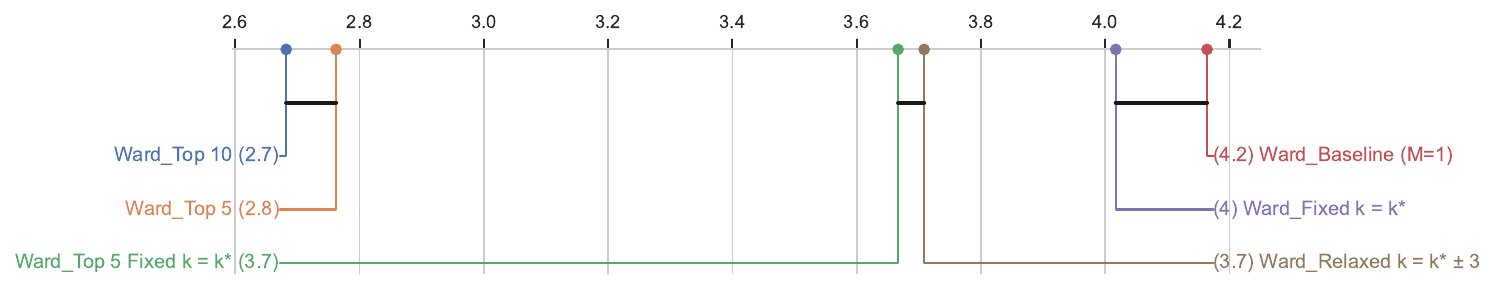}
	\caption{Critical difference diagram comparing the six FOSC-X cluster extraction scenarios using Ward-Linkage on the synthetic datasets. 
    Methods are ranked by AMI and ranks are then averaged across datasets, with groups connected by lines indicating no statistically significant difference.
    }
	\label{fig:CD_ward}
\end{figure}

\begin{figure}[!h]
	\centering
	\includegraphics[width=\textwidth, clip]{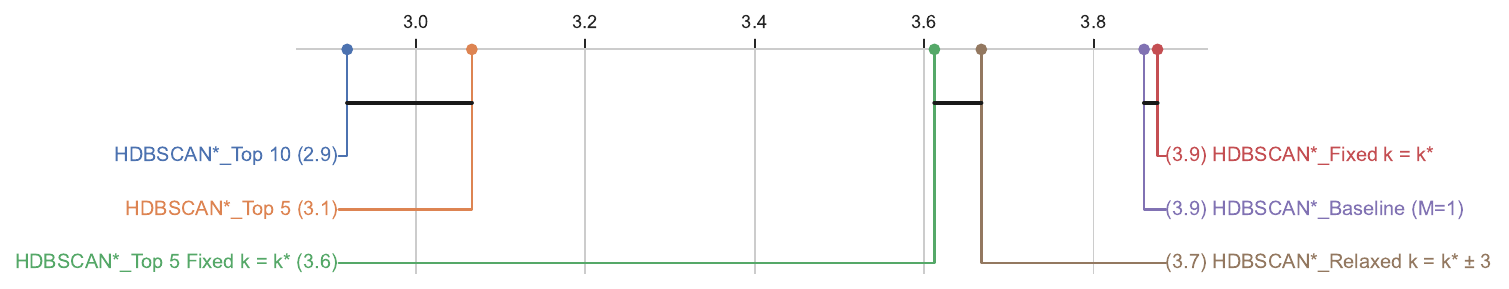}
	\caption{Critical difference diagram comparing the six FOSC-X cluster extraction scenarios using HDBSCAN* on the synthetic datasets.
    Methods are ranked by AMI and ranks are then averaged across datasets, with groups connected by lines indicating no statistically significant difference.
    }
	\label{fig:CD_hdbscan}
\end{figure}

Figure \ref{fig:box_synthetic} provides an overview of AMI performance across methods on the synthetic datasets, while the critical difference diagrams in Figures \ref{fig:CD_ward} and \ref{fig:CD_hdbscan} summarise statistical comparisons using the Friedman and Nemenyi tests at a 5\% significance level.

When interpreting the results, it is worth keeping it in mind that while by definition no configuration can produce solutions with strictly higher objective function (\emph{Stability}) than the Baseline (top-1 unconstrained) solution, the same does not hold true in terms of external agreement (AMI), which is an independent criterion.  

For Ward-Linkage, it can be seen from Figure~\ref{fig:box_synthetic}~(a) that the Baseline ($M=1$, unconstrained) exhibits the largest variability and includes several low-performing instances. All other methods generally improve upon the Baseline, with unconstrained top-$M$ achieving the strongest overall performance and typically higher AMI values. 

For HDBSCAN*, performance is higher and more consistent across all methods.\footnote{To improve interpretability, notice that the y-axis in Fig.~\ref{fig:box_synthetic}~(b) is limited within the range $[0.7,1.0]$.} As there is less room for improvements, differences relative to the Baseline are smaller, and constraining the number of clusters does not clearly provide visible gains in terms of the distribution of AMI values in this case.
In particular, the Fixed $k$ variant exhibits slightly worse performance than the Baseline and other variants. The unconstrained top-$M$ again achieves the best overall results, with slightly higher AMI values than the Baseline and other variants. 

For both algorithms, simple visual inspection of the unconstrained top-$M$ results among themselves and against the Baseline indicates that in most datasets the solution with highest AMI within the top-10 solutions can actually be found among the top-5 solutions, but in many cases it does not coincide with the top-1 (Baseline) solution.

In terms of statistical testing, the critical difference diagrams for both Ward-Linkage and HDBSCAN* (Figures~\ref{fig:CD_ward} and \ref{fig:CD_hdbscan}) exhibit a consistent pattern, identifying three statistically distinct groups of methods. The unconstrained top-$M$ variants form the highest-ranked group, while Relaxed $k$ and top-$M$ with Fixed $k$ form a second group. The Baseline and Fixed $k$ configurations form the lowest-ranked group and are not statistically distinguishable from one another, whereas all remaining methods belong to statistically higher-ranked groups.

Overall, the experiments demonstrate that exploring a small collection of multiple top-$M$ solutions can retrieve substantially better clustering alternatives in terms of an independent evaluation criterion, as compared to the single-solution baseline produced by the original FOSC. In this context, the largest gains were observed when moving from a single solution to the top-5 candidates. Increasing the number of retained solutions from top-5 to top-10 produced smaller additional improvements. 
Incorporating cluster-count constraints through $k_{\min}$ and $k_{\max}$ also provided improvements over the original baseline FOSC formulation in most cases, with a noticeable exception of the strict Fixed $k$ configuration. This suggests that while moderate granularity guidance can be beneficial also in terms of quality of results (in addition to the indisputable extra level of control offered to the user), excessive constraining may limit too much the search space and risks ultimately not being effective, especially when combined with a strict selection of the single top-1 solution. 

\subsection{Empirical runtime analysis} \label{sec:empiricalComp}

As the theoretical asymptotic computational complexity previously described in Section~\ref{subsec:complexity} represents a worst-case scenario that does not necessarily reflect typical or practical performance, we now perform an empirical evaluation\footnote{All experiments were conducted on a machine running Windows~11 with an AMD Ryzen~9 9955HX 16-Core Processor (2.50\,GHz) and 32\,GB RAM.}. A synthetic dataset with 5000 observations, 4 dimensions, and 10 ground-truth clusters is generated and clustered using Ward-Linkage to produce the cluster tree. To evaluate scaling with respect to dataset size $n$, smaller datasets are obtained by sub-sampling from this dataset to preserve a comparable hierarchical structure.

\begin{figure}[!h]
	\centering
	\includegraphics[width=\textwidth, clip]{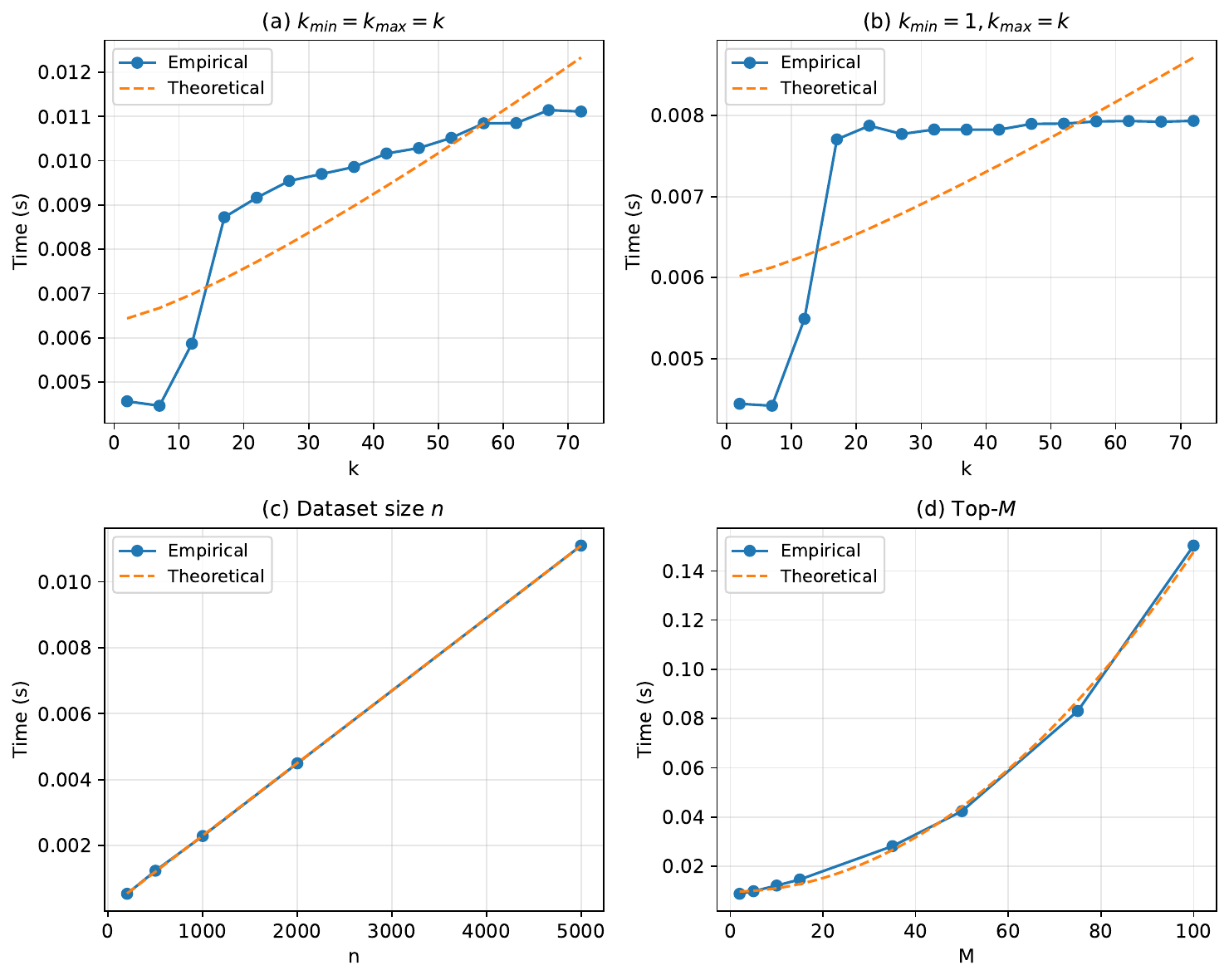}
	\caption{Empirical runtime of FOSC-X compared with the theoretical complexity $O(nKM^2 \log(MK))$ under different parameter settings. (a) Scaling with $k$ when $k_{\min}=k_{\max}=k$. (b) Scaling with $k$ when $k_{\min}=1$ and $k_{\max}=k$. (c) Scaling with the dataset size $n$. (d) Scaling with the number of tracked top-$M$ solutions. Theoretical curves are scaled by a least-squares fit to the empirical data.}
	\label{fig:comp}
\end{figure}


Figure~\ref{fig:comp} compares the empirical runtime with the theoretical complexity of $O(nk_{\max}M^2 \log(Mk_{\max}))$, where theoretical curves are scaled using least-squares regression with an intercept term to account for fixed implementation overhead and implicit constant factors. Panel (a) varies $k$ with $k_{\min}=k_{\max}=k$, while panel (b) varies $k$ with $k_{\min}=1$ and $k_{\max}=k$. For both (a) and (b), $n=5000$ and $M=1$ are fixed. 
Panel (c) varies the dataset size $n$, with $M=5$, $k_{\min}=2$, and $k_{\max}=25$ fixed to demonstrate behaviour in a non-trivial setting. Panel (d) varies the number $M$ of traced top-$M$ solutions, with $n=5000$, $k_{\min}=1$, and $k_{\max}$ unconstrained.

The empirical results show different scaling behaviour depending on the parameter varied. For both cases where $k$ is varied (Figures~\ref{fig:comp}(a) and \ref{fig:comp}(b)), the runtime grows only slowly with increasing $k$, exhibiting sub-linear behaviour that approaches near-constant scaling for larger values of $k$. After an initial increase for small $k$, the runtime stabilises, indicating that in practice the cost of increasing $k$ is minimal. This contrasts with the theoretical worst-case bound and reflects the effect of dominance pruning, which limits the number of candidate expansions in typical hierarchies. In contrast, the empirical scaling with respect to the dataset size $n$ (Figure~\ref{fig:comp}(c)) closely follows the theoretical prediction, exhibiting approximately linear growth. Similarly, when varying the number of top-$M$ solutions (Figure~\ref{fig:comp}(d)), the runtime increases quadratically, closely matching the expected $O(M^2log(M))$ behaviour.

\begin{figure}[!h]
	\centering
	\includegraphics[width=\textwidth, clip]{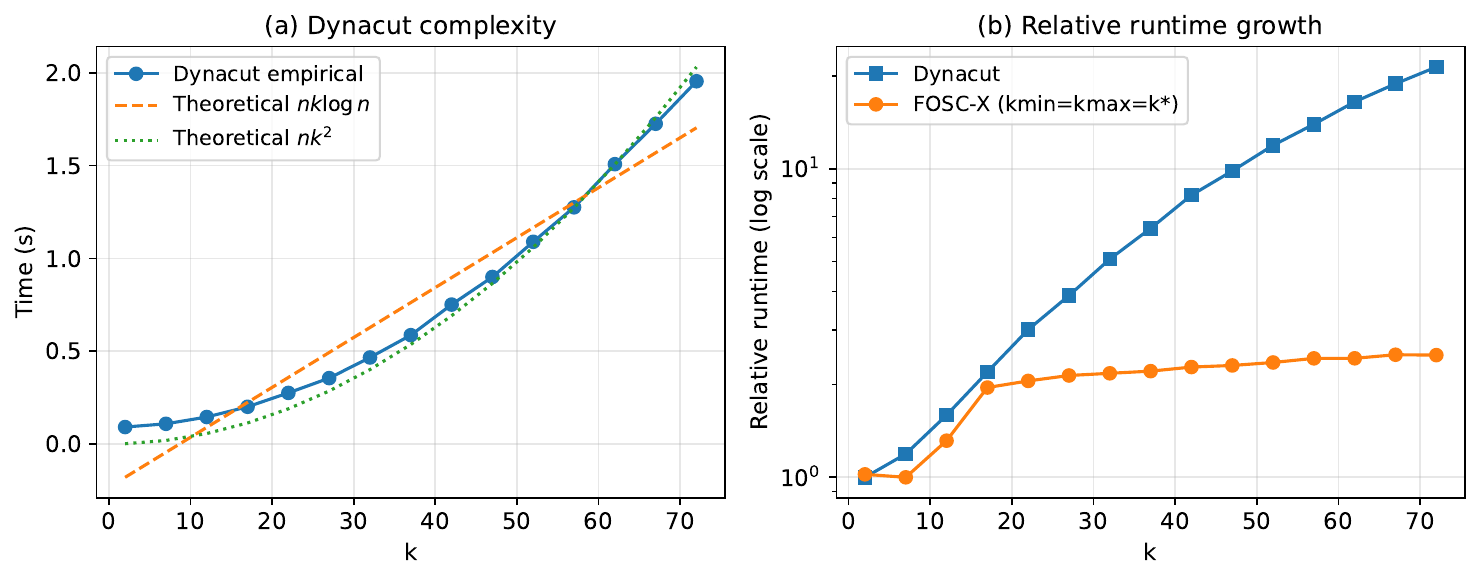}
	\caption{Empirical runtime comparison between DynaCut and FOSC-X. (a) Runtime of DynaCut as $k$ varies, shown alongside theoretical reference curves. (b) Relative runtime growth comparison between DynaCut and FOSC-X in the equivalent setting where $k_{\min}=k_{\max}$, with runtime normalised relative to the smallest evaluated value of $k$ and shown on a logarithmic scale. Theoretical curves are scaled by a least-squares fit to the empirical data.}
	\label{fig:dyna}
\end{figure}

Additionally, Figure~\ref{fig:dyna} shows (a) the empirical complexity of DynaCut~\cite{marti2017cut} as $k$ varies, and (b) a relative comparison with FOSC-X in the equivalent use case where $k_{\min}=k_{\max}$, with runtime normalised relative to the smallest evaluated value of $k$ and displayed on a logarithmic scale.\footnote{Without normalisation, our implementation of FOSC-X is orders of magnitude faster than the DynaCut implementation provided by its author in terms of absolute runtime.} Although DynaCut is reported to have complexity $O(nk\log n)$, the provided implementation exhibits approximately quadratic scaling in $k$. In this setting, FOSC-X not only provides greater control over the number of clusters through the $(k_{\min}, k_{\max})$ parameters, but also exhibits improved efficiency in the equivalent case where $k_{\min}=k_{\max}$.

Since DynaCut implicitly computes all solutions for cluster counts up to the specified value $k$, the same runtime behaviour also applies when comparing against the Relaxed $k$ setting of FOSC-X. Consistent with Figure~\ref{fig:comp}(b), relaxing the cluster-count bounds in FOSC-X results in runtime that is approximately equivalent to, and in some cases slightly lower than, the exact $k_{\min}=k_{\max}$ setting.

\section{Conclusions} \label{sec:Conclusion}

In this work, we introduced FOSC-X, an extension of the FOSC framework that enables the extraction of multiple high-quality flat clusterings from a hierarchical tree. Unlike existing approaches, FOSC-X supports the enumeration of the top-$M$ solutions and explicit constraints on the number of clusters through user-specified bounds $k_{\min}$ and $k_{\max}$, both independently as well as combined.

To achieve this, we developed a dynamic programming procedure that combines objective-based ranking with feasibility-aware dominance pruning. This allows the algorithm to efficiently explore the space of candidate clusterings while promptly discarding partial candidate solutions as soon as they are identified as being unable to contribute to any feasible global optimum.

We further showed that, although incorporating cluster-count constraints introduces additional combinatorial hurdles, dominance-based pruning ensures that the resulting procedure retains linear runtime complexity with the size $N$ of the cluster tree, which also translates into linear complexity with respect to the dataset size $n$.

FOSC-X generalises two existing extraction methods, namely, FOSC and DynaCut. Any flat clustering produced by these methods can also be obtained within the \mbox{FOSC-X} framework. In the case of FOSC, the corresponding solutions are recovered with the same computational complexity, while for DynaCut the equivalent solutions can be obtained with improved empirical efficiency. Thus, FOSC-X provides a strictly more flexible framework while subsuming the behaviour of these existing approaches.

Since the globally optimal solution is necessarily contained within the top-$M$ set, FOSC-X is never worse than standard single-solution extraction. However, by also returning an augmented collection with near-optimal solutions, FOSC-X can reveal meaningful alternatives and structural variability within the cluster tree. Through experimental evaluation, we observe that alternative high-quality solutions often exist and may provide improved interpretability or robustness.

The introduction of $[k_{\min}$,$k_{\max}]$ constraints further enables controlled extraction of cluster structure. By limiting the maximum number of selected clusters, the method can also suppress insignificant branches and reduce the influence of noise. In practice, this may yield useful solutions even when the underlying cluster tree is imperfect, for example due to suboptimal parameter choices or algorithmic mismatch. In the experimental evaluation it was found a relaxed constraint often provides better results compared to a strict setting.

A limitation of FOSC-X, which is shared with other flat clustering extraction approaches from hierarchies (see related work in Section~\ref{sec:Background}), is that it operates on a fixed, preprocessed cluster tree. The quality of the extracted flat clusterings is therefore fundamentally bounded by the structure encoded in the tree itself. If the tree does not reflect meaningful structure in the data, no extraction procedure can recover it.

The FOSC-X methodology guarantees optimality with respect to the chosen additive quality measure. The practical effectiveness of the extracted flat clusterings depends on how well the selected quality function captures meaningful structure in the data. At present, only a limited number of quality measures have been developed for use within the FOSC (and now also FOSC-X) framework, and their suitability may vary across problem domains and clustering algorithms. Future work should therefore focus on the development of additional additive quality measures and on a systematic investigation of their behaviour under different data characteristics and clustering paradigms. Expanding the space of compatible objective functions would further enhance the flexibility and applicability of optimisation based cluster extraction.

\section*{Funding}
This work was supported by the Novo Nordisk Foundation (grant no. NNF23OC0079660).

\newpage

\bibliography{FoscBib}

\newpage
\appendix

\newpage
\section{Complexity Analysis} \label{App_B} 

\subsection{Complexity under $k_{\min}$--$k_{\max}$ constraints} 
\label{sec:complexity}

We analyse the worst-case time complexity of FOSC-X under the cluster-count constraint $k_{\min} \le |z_{C_i}| \le k_{\max}$. The analysis is conducted under two assumptions: (i) the cluster tree is perfectly balanced, and (ii) the tree is binary.

The balanced tree assumption captures the worst-case setting, as the number of candidate solutions processed at a node is maximised when all child subtrees are large and of comparable size. The binary assumption is made for simplicity, and is without loss of generality, since any non-binary tree can be represented as a sequence of binary merges.

Our goal is to determine how the running time scales with the dataset size $n$ (equivalently, the number of leaves in the hierarchy) and with the upper bound $k_{\max}$. The analysis proceeds in three steps: (i) determine how many partial clusterings a node can propagate upward, (ii) compute the cost of merging two children at a fixed depth $d$, and (iii) sum this cost over all depths of the tree.

Under the dominance pruning rule discussed earlier (Section~\ref{sec:upper_bound}), at most $M$ partial clusterings are retained for each possible cluster cardinality. Since the number of clusters cannot exceed $k_{\max}$, each node retains at most $Mk_{\max}$ partial clusterings. For simplicity, we first analyse the case $M=1$; the general case introduces only a multiplicative factor of $M$.

\subsection*{Subtree sizes as a function of depth}

For a perfectly balanced binary tree of height $H$, we have
\begin{equation}
n = 2^H.
\label{eq:N_equals_2H}
\end{equation}
At exactly depth $d$ from the root, there are $2^d$ nodes, each corresponding to a subtree containing $n/2^d$ leaves. The child subtrees of these nodes therefore contain
\begin{equation}
n_d = \frac{n}{2^{d+1}}
\label{eq:subtree_size}
\end{equation}
leaves. Since the cost of processing a node depends on the sizes of its child subtrees, the complexity depends on $d+1$. Equation~\eqref{eq:subtree_size} follows from the fact that in a balanced binary tree, each increase in depth halves the number of leaves contained within each child subtree, starting from the root subtree of size $n$. The cost of processing a node depends on the sizes of its child subtrees, hence the dependence on $d+1$.

\subsection*{Number of partial clusterings per node}

A subtree containing $n_d$ leaves cannot produce more than $n_d$ clusters, since each cluster must contain at least one leaf. On the other hand, the global constraint ensures that no node can retain more than $k_{\max}$ cluster counts. Therefore, the maximum number of partial clusterings propagated from a node at depth $d$ is bounded by $\min(k_{\max}, n_d)$.

To determine where the limiting factor changes from $k_{\max}$ to $n_d$, we solve
\begin{equation}
\frac{n}{2^{d+1}} = k_{\max}.
\label{eq:threshold_condition}
\end{equation}
Solving~\eqref{eq:threshold_condition} yields
\begin{equation}
D = \left\lfloor \log_2\!\left(\frac{n}{k_{\max}}\right) \right\rfloor - 1.
\label{eq:D_definition}
\end{equation}
For $0 \le d \le D$, we have $n_d \ge k_{\max}$ and the global constraint is active. For $d > D$, we have $n_d < k_{\max}$ and subtree size becomes the limiting factor.

\subsection*{Cost of merging two children}

At depth $d$, each node receives at most $\min(k_{\max}, n_d)$ partial clusterings from each child. The Cartesian product therefore produces at most $\min(k_{\max}, n_d)^2$ candidates. Sorting these candidates prior to dominance pruning requires
\begin{equation}
O\!\left(\min(k_{\max}, n_d)^2 \log \min(k_{\max}, n_d)\right)
\label{eq:sorting_cost}
\end{equation}
time per node.

Since there are $2^d$ nodes at depth $d$, and leaf nodes at depth $H$ do not require further processing, the total running time is
\begin{equation}
T(N,k_{\max}) = \sum_{d=0}^{H-1} 2^d \, \min(k_{\max}, n_d)^2 \log \min(k_{\max}, n_d).
\label{eq:total_sum}
\end{equation}
Splitting the sum at depth $D$ gives
\begin{equation}
T(N,k_{\max}) = T_1 + T_2,
\label{eq:T_split}
\end{equation}
where
\begin{equation}
T_1 = \sum_{d=0}^{D} 2^d \, k_{\max}^2 \log k_{\max},
\label{eq:T1_def}
\end{equation}
and
\begin{equation}
T_2 = \sum_{d=D+1}^{H-1} 2^d \, n_d^2 \log n_d.
\label{eq:T2_def}
\end{equation}

\subsection*{Upper region: $0 \le d \le D$}

Bringing constants outside the sum in~\eqref{eq:T1_def} gives
\begin{equation}
T_1 = k_{\max}^2 \log k_{\max} \sum_{d=0}^{D} 2^d.
\label{eq:T1_sum}
\end{equation}
Using the geometric identity $\sum_{d=0}^{D} 2^d = 2^{D+1}-1$,
\begin{equation}
T_1 = k_{\max}^2 \log k_{\max} (2^{D+1}-1).
\label{eq:T1_closed}
\end{equation}
From~\eqref{eq:D_definition}, we have
\[
2^{D+1}
=
2^{\left\lfloor \log_2\!\left(\frac{n}{k_{\max}}\right) \right\rfloor}.
\]
Since the floor operation can reduce the exponent by at most one, it follows that
\[
\frac{1}{2}\frac{n}{k_{\max}}
\le
2^{D+1}
\le
\frac{n}{k_{\max}}.
\]
Thus replacing $2^{D+1}$ by $\frac{n}{k_{\max}}$ affects only constant factors and does not alter the asymptotic complexity. Substituting into~\eqref{eq:T1_closed} yields
\begin{equation}
T_1 = k_{\max}^2 \log k_{\max} \left(\frac{n}{k_{\max}} - 1\right) = n k_{\max} \log k_{\max} - k_{\max}^2 \log k_{\max}.
\label{eq:T1_explicit}
\end{equation}

\subsection*{Lower region: $d > D$}

Substituting $n_d = \frac{n}{2^{d+1}}$ into~\eqref{eq:T2_def} gives
\begin{equation}
T_2 = \sum_{d=D+1}^{H-1} 2^d \left(\frac{n}{2^{d+1}}\right)^2 \log\!\left(\frac{n}{2^{d+1}}\right).
\label{eq:T2_sub}
\end{equation}
Since $\frac{n}{2^{d+1}} \le k_{\max}$ for $d>D$, we bound $\log\!\left(\frac{n}{2^{d+1}}\right) \le \log k_{\max}$. Simplifying the quadratic term gives $2^d \left(\frac{n}{2^{d+1}}\right)^2 = \frac{n^2}{2^{d+2}}$, and therefore
\begin{equation}
T_2 = n^2 \log k_{\max} \sum_{d=D+1}^{H-1} 2^{-(d+2)}.
\label{eq:T2_geometric}
\end{equation}
Using the finite geometric-series identity
\begin{equation}
    \sum_{i=0}^{m-1} r^i = \frac{1-r^m}{1-r},
\end{equation}

with \(r=\tfrac12\), we obtain
\begin{equation}
\sum_{d=D+1}^{H-1} 2^{-(d+2)}
=
2^{-(D+3)}
\sum_{i=0}^{H-D-2}
\left(\frac12\right)^i
=
2^{-(D+2)}
\left(
1-2^{D+1-H}
\right).
\label{eq:geom_eval}
\end{equation}
Using $n=2^H$ and $2^{D+1} = \frac{n}{k_{\max}}$, we obtain $2^{D+1-H} = \frac{1}{k_{\max}}$ and $2^{-(D+2)} = \frac{k_{\max}}{2n}$. Substituting gives
\begin{equation}
T_2 = \frac{n k_{\max}}{2} \left(1 - \frac{1}{k_{\max}}\right) \log k_{\max} = \frac{n k_{\max}}{2} \log k_{\max} - \frac{n}{2} \log k_{\max}.
\label{eq:T2_explicit}
\end{equation}

\subsection*{Final complexity}

Combining~\eqref{eq:T1_explicit} and~\eqref{eq:T2_explicit}, we obtain the explicit expression
\begin{equation}
T(N,k_{\max}) = \frac{3}{2} n k_{\max} \log k_{\max} - k_{\max}^2 \log k_{\max} - \frac{n}{2} \log k_{\max}.
\label{eq:T_final_explicit}
\end{equation}
The dominant term is proportional to $n k_{\max} \log k_{\max}$, and therefore
\begin{equation}
T(n,k_{\max}) = O(n k_{\max} \log k_{\max}).
\label{eq:T_final_asymptotic}
\end{equation}

\subsection{Top-$M$ complexity}

For any fixed $M \ll n$, the number of feasible partial clusterings of a subtree grows rapidly with subtree size. Consequently, except near the leaves where the number of feasible partial clusterings is inherently limited, each internal node retains exactly $M$ solutions after dominance pruning. When two children each propagate $M$ partial clusterings, the merge step produces $M^2$ candidate combinations. Sorting these candidates requires $O(M^2 \log M)$ time per node, yielding an overall complexity of
\begin{equation}
T(n,M)
=
O(nM^2\log M).
\end{equation}

\paragraph{Cluster-count constrained top-$M$.}

Under cluster-count constraints, at most $M$ partial clusterings are retained for each admissible cluster count. Since there are at most $k_{\max}$ admissible counts, each node retains at most $Mk_{\max}$ partial clusterings. Combining this with the amortised analysis developed above for the constrained case yields

\begin{equation}
    T(n,M,k_{\max})
=
O\!\left(
n\,M^2\,k_{\max}\,\log(Mk_{\max})
\right).
\end{equation}

\end{document}